\documentclass[journal,transmag]{IEEEtran}

\ifCLASSINFOpdf
\usepackage[pdftex]{graphicx}
\else
\fi

\usepackage{color,soul}
\usepackage{epstopdf}
\usepackage{subcaption}
\usepackage[table]{xcolor}

\pdfoutput=1

\makeatletter
\newcommand\footnoteref[1]{\protected@xdef\@thefnmark{\ref{#1}}\@footnotemark}
\makeatother

\begin{document}

\title{Defining Image Memorability using the Visual Memory Schema}

\author{\IEEEauthorblockN{Erdem Akagunduz\IEEEauthorrefmark{1},~\IEEEmembership{Member,~IEEE},
Adrian G. Bors\IEEEauthorrefmark{1},~\IEEEmembership{Senior Member,~IEEE},
Karla K. Evans\IEEEauthorrefmark{2}\\}
\IEEEauthorblockA{\IEEEauthorrefmark{1}Department of Computer Science, University of York, UK\\}
\IEEEauthorblockA{\IEEEauthorrefmark{2}Department of Psychology, University of York, UK}
}

\markboth{}%
{Shell \MakeLowercase{\textit{et al.}}}

\IEEEtitleabstractindextext{
\begin{abstract}
Memorability of an image is a characteristic determined by the human observers' ability to remember images they have seen. Yet recent work on image memorability defines it as an intrinsic property that can be obtained independent of the observer. {The current study aims to enhance our understanding and prediction of image memorability, improving upon existing approaches by incorporating the properties of cumulative human annotations.} We propose a new concept called the Visual Memory Schema (VMS) referring to an organization of image components human observers share when encoding and recognizing images. The concept of VMS is operationalised by asking human observers to define memorable regions of images they were asked to remember during an episodic memory test. We then statistically assess the consistency of VMSs across observers for either correctly or incorrectly recognised images. The associations of the VMSs with eye fixations and saliency are analysed separately as well. Lastly, we adapt various deep learning architectures for the reconstruction and prediction of memorable regions in images and analyse the results when using transfer learning at the outputs of different convolutional network layers. 
\end{abstract}

}

\maketitle
\IEEEdisplaynontitleabstractindextext
\IEEEpeerreviewmaketitle

\section{Introduction}
\IEEEPARstart{M}{emories} are an essential component of how we define ourselves and play a crucial role in learning \cite{Tulving1972}. {There are studies that argue for a massive capacity of human episodic memory for visual information} \cite{Brady2011,Standing1975}. The study of human memory capacity for visual information such as complex images has sparked interest in a number of different scientific fields not only in psychology but in computational intelligence, as well \cite{Isola2011a,Isola2011b,Bylinskii2015,Khosla2012a,Oliva2001}. Understanding the human ability to remember information from images has a significant impact on furthering our knowledge about the human mind, for the  development of new technologies in mental augmentation, information retrieval and marketing just to name a few.

Within the last decade, there has been a growing interest in understanding the memorability of an image as an intrinsic property of the image itself. A pioneering study by Isola et al. \cite{Isola2011a} found a high consistency among observers as to which images were best remembered and demonstrated that computer vision techniques allowed for good prediction of image memorability. There have been other studies that have related intrinsic image memorability to attribute annotations \cite{Isola2011b}, object annotations \cite{Kim2013, Isola2014}, automatic semantics \cite{Khosla2012a}, visual attention \cite{Mancas2013, Bylinskii2015}, saliency \cite{Mancas2013} or image category information \cite{Bylinskii2015}. Recently, Khosla et al. obtained memorability scores using Mechanical Turk for a large image set and achieved high prediction rates by using deep neural networks \cite{Khosla2015}. All of the aforementioned studies collected data using the same experimental methodology, in which participants view a sequence of images and are asked to respond whenever they see an identical repeat of an image at any time in the sequence. The aim of their experiments is to measure the memorability of the image as a global property, independent of the relations among the local regions of the image. Although there are studies that focus on region-based memorability \cite{Khosla2012a} and use different experimental designs for this purpose \cite{Dubey2015} (such as showing pre-segmented parts of images instead of identical repeats), none of these studies have asked the human observers to indicate the regions that made them remember the images.

\begin{figure}[t]
\centering
\begin{subfigure}{1.9cm}	
	\includegraphics*[width=1.9cm, height=1.9cm]{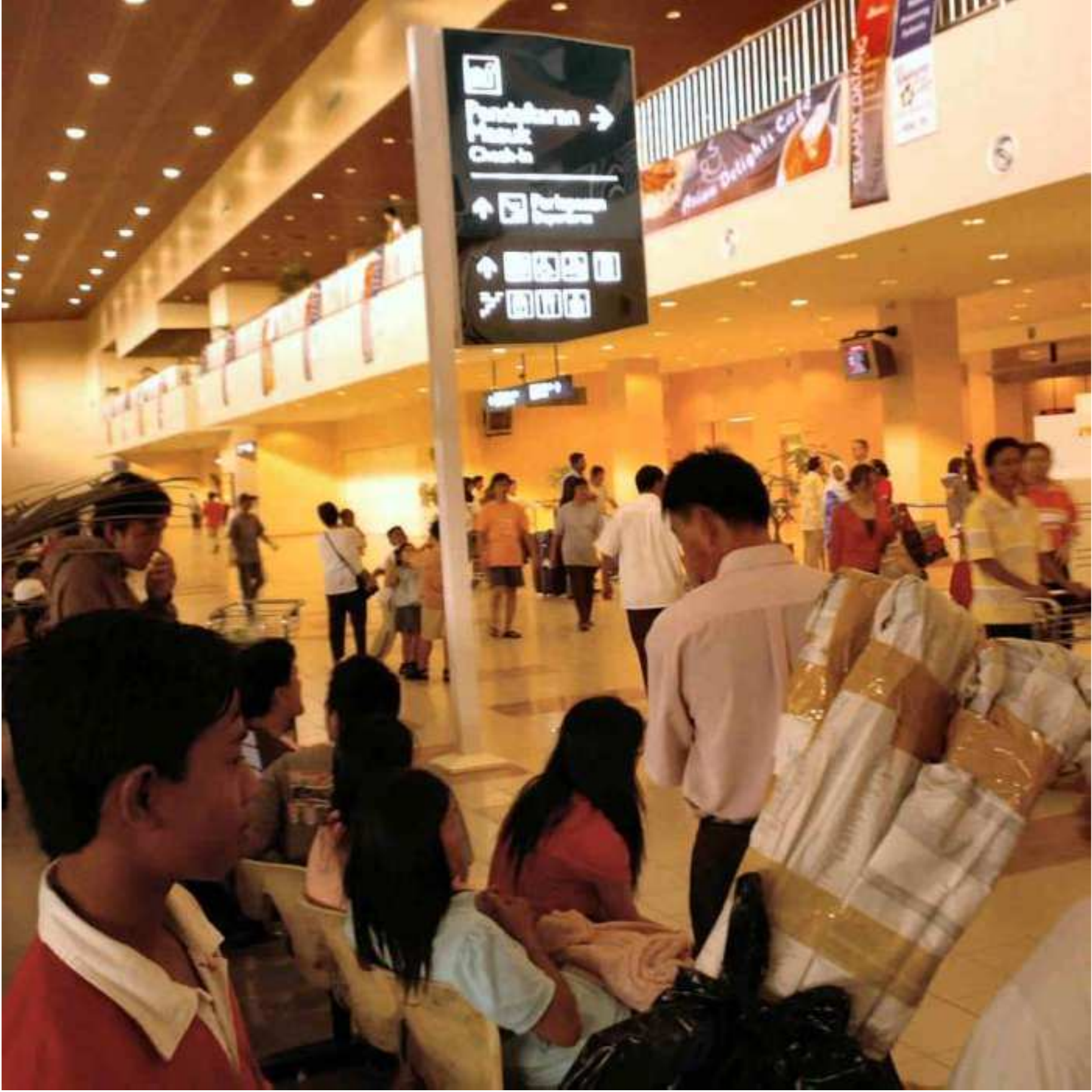}
	\caption{Image}	
\end{subfigure}
\begin{subfigure}{1.9cm}	
	\includegraphics*[trim= -10 0 10 0,clip=false,width=1.9cm, height=1.9cm]{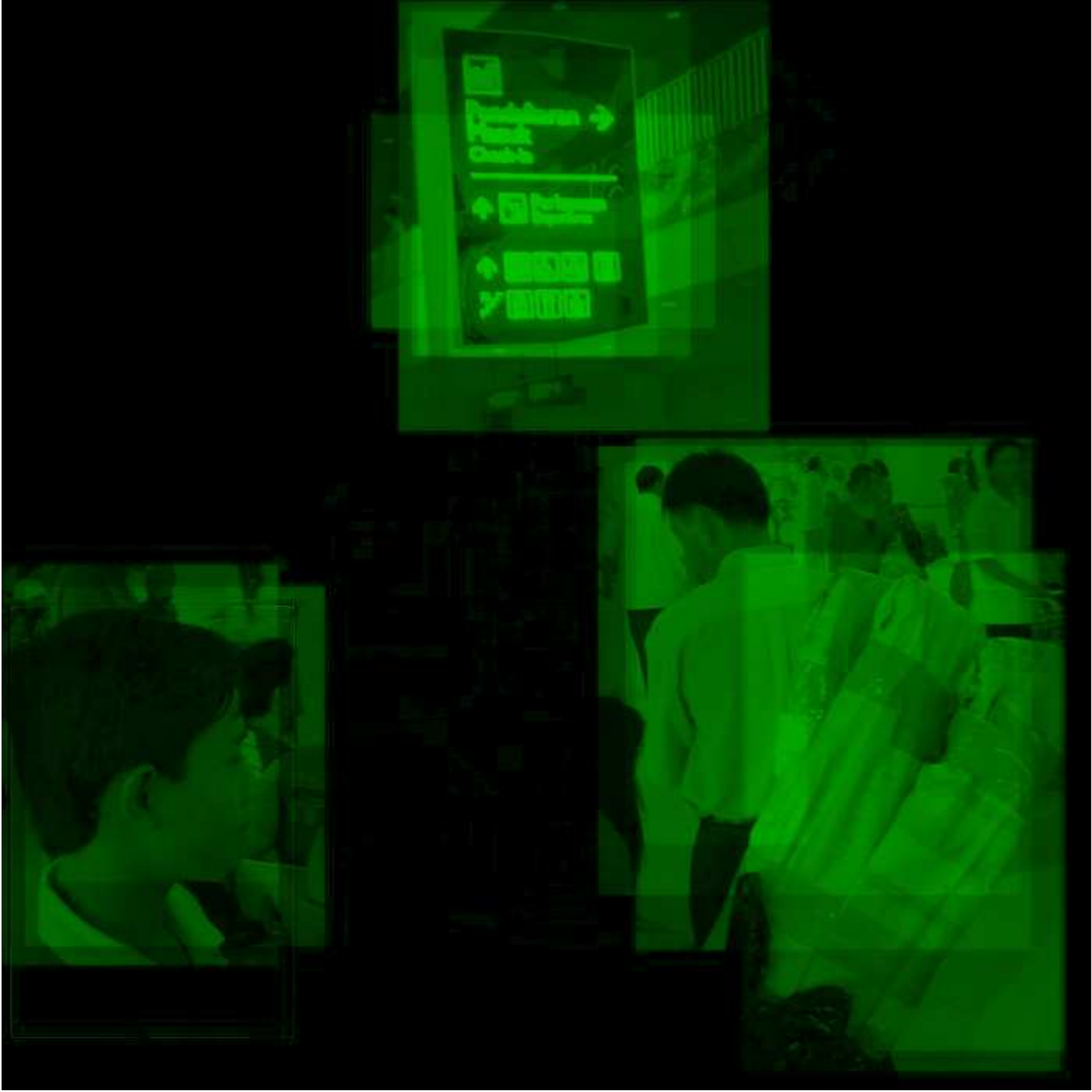}
	\caption{True VMS}	
\end{subfigure}
\begin{subfigure}{1.9cm}	
	\includegraphics*[trim= -30 0 30 0,clip=false,width=1.9cm, height=1.9cm]{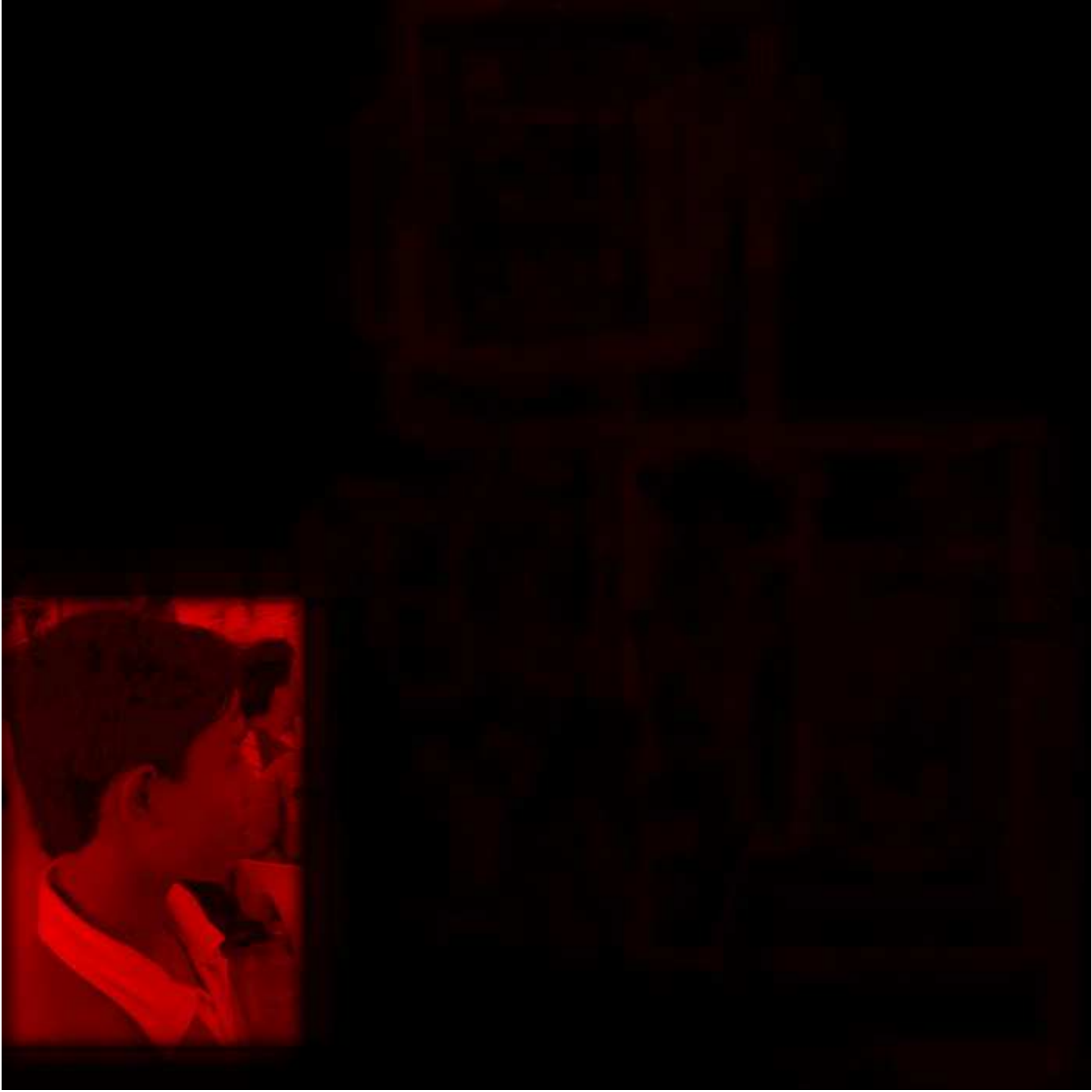}
	\caption{False VMS}	 
\end{subfigure}
\begin{subfigure}{2.8cm}	
	\includegraphics*[trim= -120 0 120 0,clip=false,width=1.9cm, height=1.9cm]{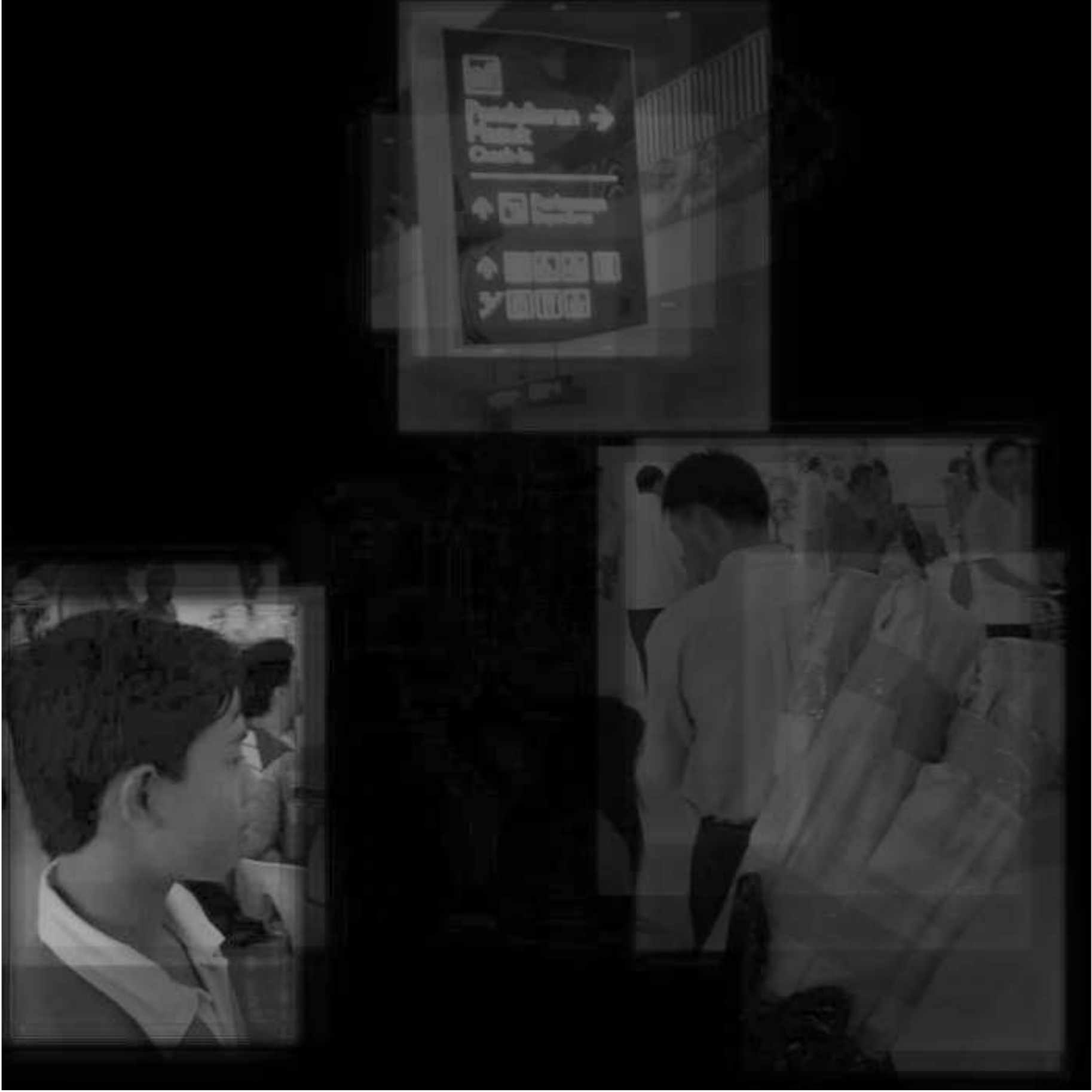}
	\caption{Combined VMS}	 
\end{subfigure}
\caption{Visual memory schemas (VMSs) corresponding to {correct (b), false (c) and both correct and false retrievals} (d) of the image shown in (a). {In this paper, visual memory schemas correspond to human-annotated regions which are pooled across observers, who are asked during a memory experiment to indicate the regions of the image that made them remember that image.}}
 \label{VMSchema}
\end{figure}

In this paper, we propose a novel approach to investigating image memorability in which we first ask 90 participants to memorize 400 images and then during a test phase rate how well they remember each of these images and select those image regions that made them remember it. Our aim is to further our understanding of how humans remember images and what they find memorable in these images. Here, we analyse statistically whether the regions, indicated by the observers as being seen before, are consistent across different groups of observers and how do they correlate with the measure of image memorability defined by {Isola} et al. \cite{Isola2011a}. In this context, we define the accumulated memorable parts of an image, selected across observers as the Visual Memory Schema (VMS), a framework of mental representation that observers use to organize their memory for future retrieval. We further define true and false VMSs to indicate whether the selected regions are from an image that is correctly recalled or from a image that was a false alarm {(i.e.  an image that the participants remember as seeing but actually has never been shown before)}. We then use machine learning techniques to estimate the memorability of an image using VMS. Finally, by using the image structures that emerge in the layers of deep convolutional neural networks \cite{Khosla2015,Simonyan15,Chatfield14}, we reconstruct the VMS of an image and compare it to the human-collected VMS.

The paper is organized as follows: Section II presents the psychological origins of the proposed visual memory schema concept and how this is operationalised. {In Section III, we provide the methodology of the memory experiment. The fourth section presents an analysis on the proposed concept compared to other concepts such as the visual saliency and the overt attention. Section V provides an assessment of whether an addition of visual memory schema enhances the power of various computer vision features, estimated from images, to predict image memorability.  Section VI presents the analysis of the reconstruction of the VMS selections in images using five different deep learning architectures, while Section VII outlines the conclusions of this study.}

\section{Visual Memory Schema}
Schema is a concept used in cognitive psychology that describes an organized pattern of thought or more specifically a mental representation of a concept \cite{Bartlett1955}. It embodies a structure for organizing information into categories and relationships among the categories. People use schemas to organize current semantic and episodic knowledge that then in turn provide a framework for future understanding. Examples of a schema include categories, stereotypes, etc. A schema can also be viewed as a tool for organizing our memories. For example when we observe an image it is not the individual pixels or their distribution in the the image that we extract in order to remember that image. But rather the visual schemas associated with the image category and composed of the key regions of the image, objects and relations between those objects that are idiosyncratic of that image \cite{Mandler1977}.

Visual schemas are represented by objects and scene regions in terms of their physical properties and the spatial arrangements of their components. More specifically, they correspond to  mental representations of how different regions of a scene and their content are related, organized and encoded into visual memory. Memories organized in this manner are efficient and allow for successful retrieval when a scene is seen again \cite{Mandler1977}. {However, visual memory schemas, we hypothesize, can also bring about proactive interference for information observed in images with the previously accumulated information}. This proactive interference \cite{Anderson1996} may lead to false memories resulting in false alarms upon retrieval. For example certain parts of an image being seen by a person, who actually has never seen that image before, may resemble visual schemas related to past life experiences. What is more, past experiments from psychology on human long-term memory \cite{Vogt2005} clearly show that humans are very bad in their memory for pure texture without any semantics attached to it or image sets of only homogeneous exemplars from one semantic scene category. We hypothesize that the role of semantics organisation, i.e. \emph{visual memory schemas} is critical for memorability of images, yet still unrecognised.

A Visual Memory Schema (VMS) in our experiment was defined individually for each image as a map of visual regions that are likely to be remembered from that image. VMS is not just another map that depicts the regional memorability strength of an image, but the organization of the visual schemas defined by observers themselves. It carries knowledge of both what the observers truly have encoded as well as what they think an image should contain based on their semantic knowledge and episodic experience of the world. People may incorrectly think that they have seen an image, and may recall regions that made them think they remember. Thus, there are VMSs corresponding to both true and false image selections, which can be assessed through a visual memory experiment, as proposed in this research study.

In order to operationalise and define a VMS for images, we used a standard episodic memory test paradigm \cite{Tulving1972} and added a novel component in which the participants are asked to select what parts of the image made them remember the image they were asked to memorize. Then, VMS is constructed for each image using the accumulated human annotations of the image regions that represent these memorable regions as shown in Figure \ref{VMSchema}. VMS is different from the memorability map concept introduced in \cite{Khosla2012a}, in the sense that it is not a computed map but an actual ground truth of human visual memory as indicated by the participants in the visual memory experiment. {Since the human annotations may be actually correct or false (e.g. regions identified in images correctly or incorrectly recognized as seen before), we define visual memory schemas for both true and false image recognition.} The annotations obtained for an image which observers correctly remembered are accumulated to construct the \emph{true VMS}, whereas the annotations obtained for an image which observers were mistaken to think that they have seen in the image, are accumulated to construct the \emph{false VMS} (Figure \ref{VMSchema}).
VMSs are single channel maps having the same resolution as the image. When constructing a VMS for an image, the human annotations are added on top of each other and are normalized by the number of participants that annotated the image. Thus, VMS is a 2D probability distribution function (PDF) of the spatial distribution of the pixels, corresponding to specific scene information as visualized in the image. VMS indicate the probability for specific image regions of being selected by an observer as memorable. In other words, the higher value (brighter) the pixels composing the VMS become, the more likely they are to be remembered by a human observer. It is important to note that the VMS is a map constructed by using human observer responses, defining most memorable regions of images, unlike the memorability maps in Khosla et al. \cite{Khosla2012a} that were based on automatic machine computations. Furthermore VMS represents both true and false memorability of a region, which provides a different and improved concept of region memorability, when compared to previous studies on the subject \cite{Khosla2012a}. 

Our main motivation for introducing the VMS concept is its critical role in image memorability. Previous work has shown that image memorability can be obtained independent of the observer and can be predicted to a degree. There is sizeable support for this conclusion \cite{Isola2011a}, however it underestimates the fact that an image is memorable only if it has cognitive elements shared by the majority of people and it is these shared cognitive elements that renders the image memorable. When image memorability is defined as an intrinsic property of an image, it refers to a low level property hidden within the image signal. {Although this pioneering approach proves to be very instructive, it may lead us to omit the fact that memorability is something that humans bring to the image. It is after all human memory and predicting it that we are interested in. We argue that VMSs are not only a statistics of signals, but they embed a collection of human contribution as well.} For this reason we believe that, in addition to analysing image memorability with signal processing techniques, a new concept that encapsulates the cognitive organization that underpins image memorability must be introduced. Although there have been previous efforts to relate the semantics of an image to memorability, {scene categories} or object labels are quite primitive in defining cognitive organization of an image when compared to the proposed VMS concept. In most cases the visual schema hidden in an image is more complex than an object label or scene category.

Thus far the the concept of visual attention is the only concept from psychology that has been invoked to characterize memorability in computer vision. {Visual attention in computer vision is approximated either as a collection of observers' eye fixations on a region, as measured by eye gaze using an eye tracker or as a saliency map calculated by specific algorithms such as bottom-up or top-down saliency} \cite{Harel2006,Itti2000}. Such measure of overt visual attention is only weakly correlated with image memorability \cite{Isola2011b}. In the following sections, we examine the correlations of VMS with other related concepts, namely eye fixations and computed saliency. We also analyse the consistency of the VMS and show that, similar to image memorability, it has consistent results across various human observers.

\section{The Image Memory Experiment}
In order to understand the visual memory schema concept within the image memorability context, we have designed a novel approach to a standard memory experiment. In this section the image stimulus set and the methodology of the experiment are described and compared with other memory experiments.

\subsection{The VISCHEMA Image Set}
In this subsection, we explain how VISCHEMA\footnote{\label{website}http://www.cs.york.ac.uk/vischema} image set, used during the memory experiments, was formed. The memory experiment was conducted using 800 images selected from the Fine-Grained Image Memorability (FIGRIM) set \cite{Bylinskii2015}. The FIGRIM image set is composed of 1754 target images (i.e. images with memorability scores indicated by human observers) from 21 different scene categories with more than 300 images of at least $700 \times 700$  pixels in resolution, selected from among the images from the SUN image set \cite{SUN2010}. A subset of target images from the FIGRIM dataset additionally includes corresponding mappings of the observers' eye-movement locations recorded during the memory test. For the FIGRIM memory experiments, 120 images representing a mix of target and filler images were presented to human observers for one second each. Both inter-category and across-category experiments were conducted, thus two separate memorability scores exist for each image \cite{Bylinskii2015}. 

In the following we organize the images used during the memory experiment in a hierarchical categorization structure based on the principles of experimentally supported psychological prototype and exemplar theories \cite{Cohen2005} that indicate how human observers categorize objects and ideas. This theory postulates that categories form part of a hierarchical structure that when applied to taxonomy has three basic levels: the supra-ordinate or higher level, the base or middle level and the subordinate or lower level. Humans remember the observed information by creating organized patterns of thought. With this new category structure, we aim at constructing {relatively balanced} category definitions which will correlate stronger with the way humans recognize, differentiate and understand images.

The VISCHEMA dataset is organized in 12 image categories as shown in Figure \ref{CategoryHRFigure}. The image categories are organized within a hierarchical structure with the higher levels in this hierarchy consisting of \emph{Indoor} and \emph{Outdoor} supra-ordinate categories. Then, at the second level, each of these categories were labelled as either \emph{Private} or \emph{Public} for the \emph{Indoor} scenes, while for the \emph{Outdoor} scenes are labelled as \emph{Man-made} and \emph{Natural}. The categorization continues with further dividing into subordinate FIGRIM/SUN categories, such as: \emph{Kitchen} (100), \emph{Living room} (100), \emph{Air terminal} (100), \emph{Conference room} (100), \emph{Amusement park} (44), \emph{Playground} (56), \emph{House} (66), \emph{Skyscraper} (34), \emph{Golf course} (58), \emph{Pasture} (42), \emph{Badlands} (47), \emph{Mountain} (53), where the numbers of images in each subcategory is indicated in the parentheses. Each leaf-category include images from one or more of the original categories of the FIGRIM/SUN image sets, with 100 images assigned to each of the 8 basic (leaf) categories. For example, the categories \emph{Badlands} and \emph{Mountains} are combined within the \emph{Isolated} category, which is a leaf of \emph{Outdoor}/\emph{Natural} scenes. Similarly, the \emph{Airport terminal} and \emph{Conference room} categories are renamed as \emph{Big} and \emph{Small}, respectively, which are self-explanatory because they refer to the contextual space, while being the leaf categories of \emph{Indoor}/\emph{Public} category.

\begin{figure*}[t]
 \centering
	\includegraphics*[width=15.01cm, height=2.7cm]{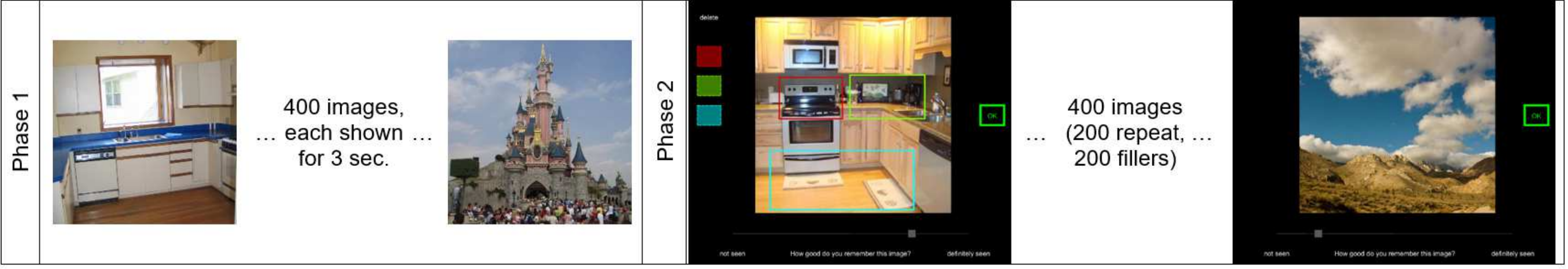}\par
	\caption{The memory experiment has two stages. During the first stage, the participants in the experiment, are shown 400 images, each for 3 seconds. During the second stage they are shown another 400 images, including 200 that are repetitions from the first stage. Participants are also asked to rate how well they think they remember the image they see and select rectangular regions from the image that made them remember it.}
 \label{ExpFigure}
\end{figure*}

For the sake of better understanding the difference between visual memory schemas of various scene categories, we avoided using certain types of images (as defined below) when selecting images from the FIGRIM set for the newly created VISCHEMA set. Previous work \cite{Isola2011a} shows that these types of images tend to dominate the composition, thus the memorability of an image. Consequently, we exclude from the VISCHEMA image set, the following types of images: images containing any kind of large text (a banner, billboard, sign that labels the image), direct shots of people posing and looking at the camera, photographic compositions of a single figure (i.e. person, animal, statue etc.), any well-known architectural structure (e.g. Empire State Building) or a well-known place (e.g. The Trafalgar square), images with a digital date in the corner, a direct shot of a flag or famous logo, or any overlaid line drawing (e.g. a curve or an arrow ). The exclusion criteria were based on findings of previous research reporting that images with the aforementioned elements, are inherently more memorable than the others, regardless of their scene category \cite{Bylinskii2015}. 

\subsection{Experimental Procedure}
For the study  90 participants were recruited from the population of students and staff at the University of York, UK (age range 19-30 years) and engaged in a memory experiment, consisting of two stages (Figure \ref{ExpFigure}). During the first stage (study phase), all participants were shown 400 images from 8 \emph{leaf} (base) categories, in a randomized order. Each image was shown for 3 seconds with the study phase of the experiment lasting a total of 20 minutes. The participants were asked to do their best to memorize the images they saw on a computer screen, in a quiet and darkened room.

The first stage was immediately followed by the second stage (test phase) in which the participants were shown another group of 400 images, 200 of which were repetitions from the first stage, in a randomized order. Similar to the first stage, the category distribution was uniform, such that 50 out of 100 images from each 8 \emph{leaf} categories were shown. During the test phase, the participants were first asked to rate how well they remembered the image using a continuous rating bar from "\emph{not seen}" to "\emph{definitely seen}". If they {thought} they {remembered} the image well enough (i.e. by placing the rating bar above the predefined threshold of 30\%) they were asked to select at least 1 and at most 3 rectangular regions, of size determined by the observer, that made them remember that image.

Each participant saw a total of 600 different images in a single experiment including 200 repeat images (images shown in the study and then again in the test phase), 200 non-repeat (first-stage-fillers) and 200 new images representing second-stage-filler images {(thus making a total 400 images at each stage)}. Each image was shown to the participants in the test phase for region selection, for approximately 45 (90 subjects $\times$ 400 second phase images / 800 total images) times across participants, ensuring an equal probability of observation for each image by the participants.

\subsection{Measuring Image Memorability}
When analysing the results of the experiment, image memorability, or \emph{hit rate} (\textbf{HR}), is defined as the proportion between the number of images, which are correctly chosen as being remembered by human observers, and the total number of their occurrences as a repeat image \cite{Isola2011a}. Similarly, the \emph{false alarm rate} (\textbf{FAR}) is the proportion of the false hits of an image from the total number of its occurrences as a second-stage-filler ({\em i.e.} non-repeat) image. In previous experiments described in the literature, a hit and a false alarm are easily determined since the participants are asked to make a yes or no decision by pressing the space bar \cite{Isola2011a,Bylinskii2015}. However in our experiments, by using an indicative bar, the participants rated their confidence in how well they {remembered} the image. Thus, the participants {were} able to express whether their response of remembering an image was vague or certain, and quantify the degree of confidence in their decision. Using a confidence scale allows us to produce ROC curves that provided us with a sensitivity measure of the experiment. {However, we had to define a threshold in the confidence values, indicated by the participants in the experiment, in order to be able to decide eventually whether the image was remembered or not}. This threshold has a direct influence on the calculation of HR and FAR values.

\begin{figure}[t]
 \centering
	\includegraphics*[trim=90 0 40 0,clip=false,width=6.8cm, height=2.8cm]{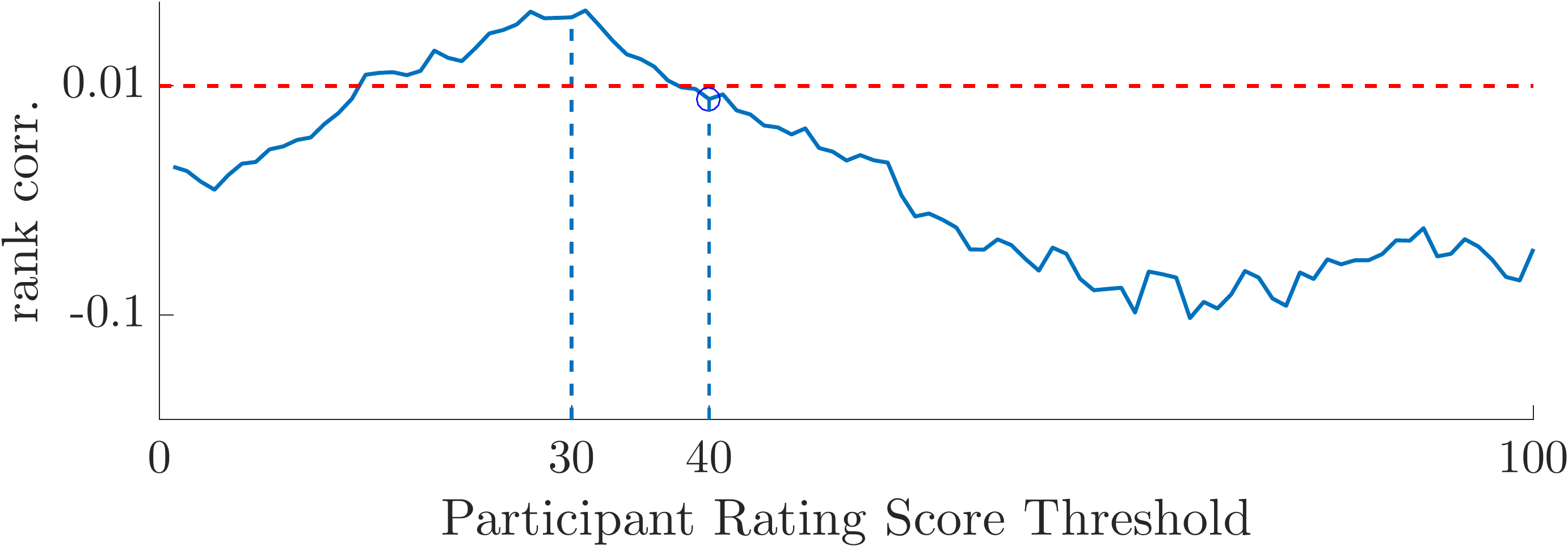}\par
	\caption{Spearman's rank correlation between the hit rates and the false alarm rates as a function of the participant rating score threshold. At the selected threshold 40, $\rho$ value is 0.0036.}
 \label{HRvsFARrho}
\end{figure}

{A range of 0-100 was used for rating the confidence in the memorisation of a specific image. A confidence value of 0 indicated strong confidence that the image was not seen before and 100 that it was definitely seen. There was a hard threshold at the level of 30. When the participants in the experiment rated above this level they were asked to select at least 1 and at most 3 regions that made them remember the image. One might argue that this is a logical threshold, indicating that the ratings above this value are the same as saying yes in the previous experiments. However when we analyse {(after we post-processed the experimental data)} how HR and FAR values change with the memorability threshold, it can be observed that the actual threshold should be around 40.} Figure \ref{HRvsFARrho} depicts the Spearman's rank correlation (\(\rho\)) between HR and FAR for any threshold value between 0 and 100. According to signal detection theory, \(\rho\) between HR and FAR must be {a positive value, very close to zero} ($<$0.01) in a natural detection scheme \cite{Wickens2001}. In other words, HR and FAR values must not increase or decrease together. Figure \ref{HRvsFARrho} shows that \(\rho\) is 0.04616 when the threshold is equal to 31, which is high for a signal detection experiment. At this threshold the participant behaviour is different from what is expected, because a participant may decide to rate below 30 because she/he does not want to select a region, although remembers seeing the image, or a participant may want to select a false region so she/he selects above 30, although she/he does not remember seeing the whole image. 

Thus, following the analysis of how HR and FAR values change according to the memorability threshold, we eventually select the value of 40 (\(\rho\)=0.0036), as the memorability threshold. This threshold was chosen because the participants in the experiment were not able to select regions when the threshold was below 30. Moreover, 40 represents the smallest threshold value for which the rank correlation \(\rho\) between HR and FAR falls below 0.01. Choosing a higher threshold would have produced HR values that are too small. Consequently, from now on all the results from this study are calculated using a confidence threshold of 40.

\begin{figure}[t]
\centering
\includegraphics*[trim=20 0 -20 0,clip=false,width=7cm, height=4cm]{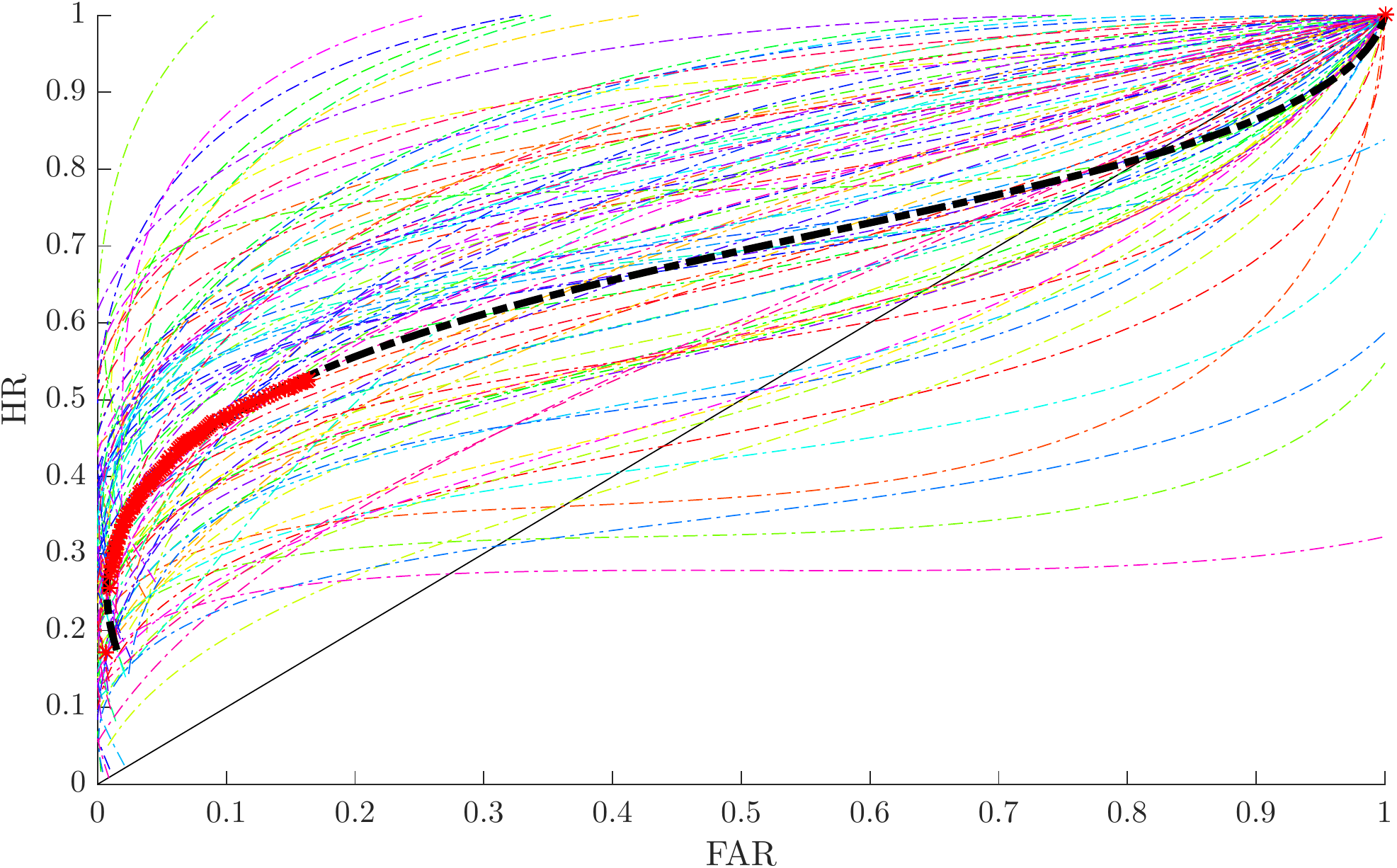}
\caption{Receiver operating characteristics (ROC) for the overall experiment {and for each participant separately. The thick-dashed line is the estimated fourth-order B\'ezier curve of the red dots, which together represent the ROC curve of the overall experiment. The thin dashed lines of various colours stand for the ROC curve results for each participant separately}.}
 \label{ROC}
\end{figure}

\subsection{Comparison with Previous Experiments}
Figure \ref{ROC} depicts the receiver operating characteristics (ROC) curve for the overall experiment and for each participant separately. As seen from Figure \ref{ROC}, even though there is a lot of expected individual variability, the ROC curve of the overall experiment results in an area-under-curve of 0.677 and sensitivity (\emph{d$^{\prime}$}) of 1.319, showing that the image memorability signal was significantly above chance and the experiment was successful.

The average hit rate of the observers in the experiment we conducted is lower than what was observed in the  FIGRIM experiments, reported in Table \ref{HRandFARAllExp}, both in the inter-category (AMT1) and across-category (AMT2) experiments. The average FAR for our experiment is also lower than that of AMT1 and AMT2, which have been reported in Bylinskii {\em et al.} \cite{Bylinskii2015}.

Considering the fact that our experiment sessions took on average 50 minutes overall (the study phase around 20 minutes and the test phase around 30 minutes) for each participant, whereas AMT1 and AMT2 took about 2 minutes each \cite{Bylinskii2015}, we would expect to see the differences between the results provided in Table \ref{HRandFARAllExp}. Moreover, in a single session we show to the participants a total of 600 images, of which 400 are different images, whereas AMT1 and AMT2 experiments used only 120 images. Thus our experiment is considerably more challenging than the previous experiments, resulting in lower observed HRs.  Furthermore, the relative difficulty and the complex methodology of our experiment compelled the participants to be more conservative when rating the memory scores, which {resulted} in lower FARs. However, the rank correlations of the HRs among different experiments show that image memorability scores {are} consistent among experiments. AMT1 and AMT2 experiments have a rank correlation of 0.594\footnote{The HRs and FARs for the AMT1 and AMT2 experiments are re-calculated only for the 800 images from the VISCHEMA Image set. For this reason these values differ from those reported in \cite{Bylinskii2015}.} while the rank correlations between our experiment and AMT1/AMT2 are on the order of 0.5028/0.54066, respectively.

\begin{figure}[t]
	\includegraphics*[trim=0 0 0 0,clip=true,width=8.8cm, height=4cm]{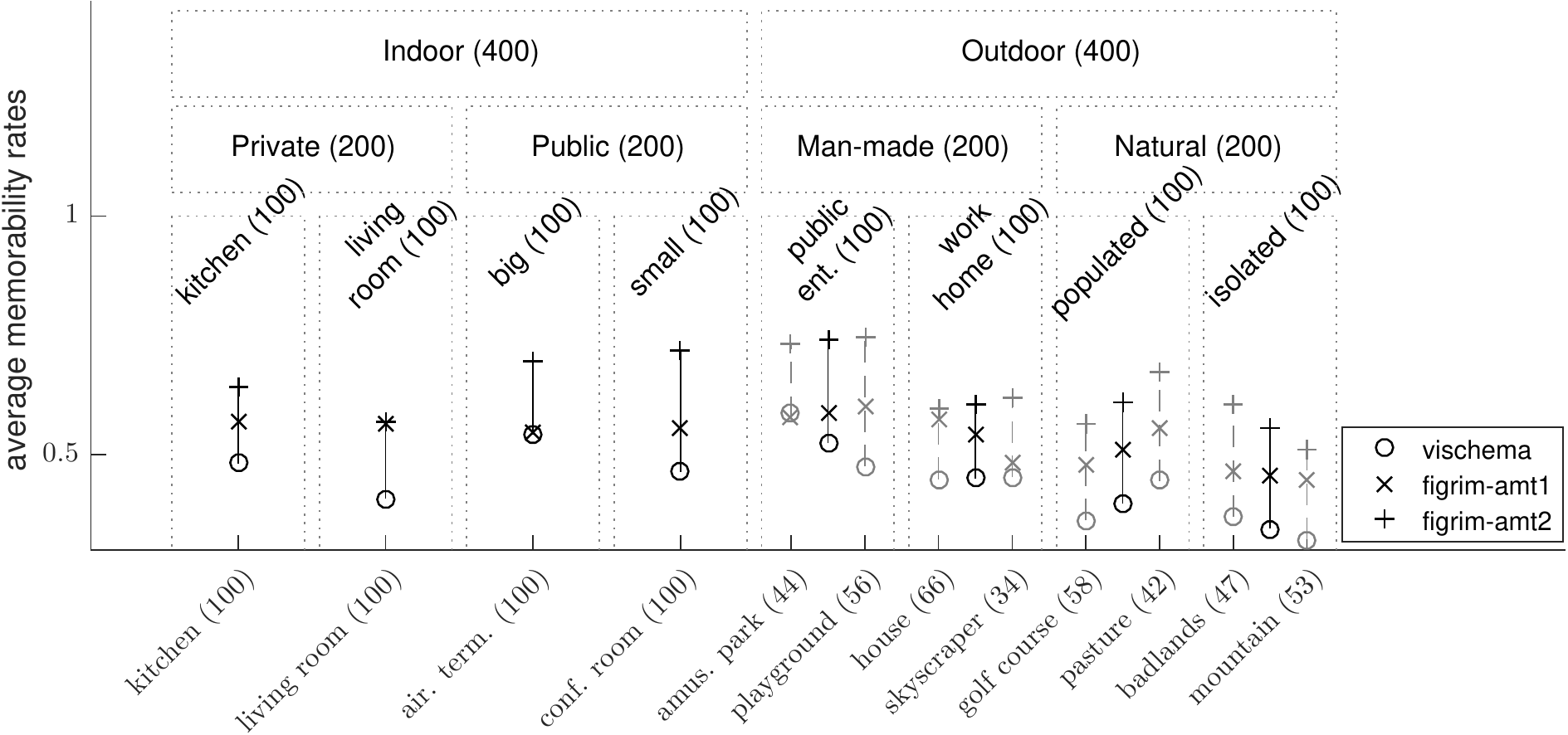}\par
	\caption{Categories of images in a hierarchical structure used for the experiments and average hit rates obtained from the three memory experiments: proposed framework indicated by circles and AMT1 \cite{Bylinskii2015} indicated by crosses and AMT2 \cite{Bylinskii2015} indicated by plus sign. The number of images used for each FIGRIM/SUN category is indicated next to the category labels on x-axis. Results are indicated for each category separately.}
 \label{CategoryHRFigure}
\end{figure}

\begin{table}[t]
\centering
\caption{Average HR and FAR values with their corresponding standard deviations compared with the results from the FIGRIM experiments \cite{Bylinskii2015}.}
\label{HRandFARAllExp}
\begin{tabular}{| c | c | c | c |}
 \hline
 \(\mu\) (\(\sigma\)) & VISCHEMA & AMT1 & AMT2 \\  \hline
HR & 0.451 ($\pm$0.175) & 0.5405 ($\pm$0.157) & 0.634 ($\pm$0.140) \\  \hline
FAR & 0.075 ($\pm$0.094) &  0.150 ($\pm$0.110) & 0.111 ($\pm$0.090) \\  \hline
\end{tabular}
\end{table}

\section{Analysing Visual Memory Schemas}
VMSs are constructed for each image by adding together all region selections  made by the participants in the experiment for that image and then normalizing the result by dividing with the number of times that the image was annotated. A VMS corresponds to the probability density function (PDF) of that decisions about their memories made for each image by all participants. The participants may annotate both repeat and second-stage filler (non-repeat) images. Therefore there are two types of selections made by the participants in the memory experiment: true or false. Consequently, we can define a true VMS and a false VMS for each of the 800 images.

A VMS represents a PDF of the image selections made by the participants in the experiment. The estimations of either visual saliency or eye fixations can also be represented as PDFs for a certain image, allowing us to compare VMS to these measures. In the following we use two well-known statistical measures for comparing two distribution functions in order to measure the relationships between VMSs with either visual saliency maps or with eye fixation maps. 

The first measure used for this comparison is the Pearson linear correlation coefficient\footnote{The Pearson's correlation coefficient, called also the normalized-cross correlation, is used to calculate the relation between the true and false VMSs with the eye fixation or with the saliency maps. It should not be confused with the Spearman's rank correlation that we use to calculate the relation between the results of two memory experiments.}, denoted as \(\rho\)$^{2D}$, which compares two 2D maps with pixel values ranging between [0,1] and of the same resolution, A and B. It is given by the following equation:

\begin{equation}
\rho^{2D}_{A,B}=\frac{1}{n}\cdot \sum_{i,j}^{ }\frac{\left (A(i,j)-\mu_{A} \right )\cdot \left ( B(i,j)-\mu_{B}  \right )}{\sigma_{A}\cdot \sigma_{B}}
\label{NCC}
\end{equation}

\noindent
where \emph{n} is the total number of pixels in A or B, \(\mu\)$_{A}$ and \(\mu\)$_{B}$ are the average pixel values and \(\sigma\)$_{A}$ and \(\sigma\)$_{B}$ are the standard deviations of the pixels of A and B, respectively. The correlation \(\rho\)$^{2D}$  is a measure of linear dependence between two maps and it ranges between \mbox{[-1,+1]} with +1 showing complete positive dependence, -1 showing complete negative dependence and 0 showing independence. 

The second measure used is the mutual information (MI) criterion, denoted as I$_{A,B}$ between the PDFs characterizing the discrete random variables A and B:

\begin{equation}
I_{A,B} = \sum_{a\epsilon B}^{ } \sum_{b\epsilon B}^{ } p(a,b)\cdot log\left (  \frac{p(a,b)}{p(a)p(b)}\right )\emph{}da\emph{ }db
\label{MIC}
\end{equation}

\noindent
where \emph{p(x,y)} is the joint probability distribution function of A and B, and \emph{p(a)} and \emph{p(b)} are the marginal probability distribution functions of A and B respectively. I$_{A,B}$ takes values in the range [0,+$\infty$).  For example, if A and B are independent of each other, then by knowing A we do not have any information about B and vice versa,  and consequently their mutual information is zero. At the other extreme, if A is a deterministic function of B (and vice versa) then all information conveyed by A is shared with B, and the mutual information is the same as the uncertainty contained in A or B alone, which is actually their entropy.

As a distance criterion, MI is more general when compared to the correlation (\(\rho\)$^{2D}$),  because the correlation only takes into account the linear relationships between two distributions whereas MI can handle non-linear relationships as well. Nevertheless, we use both criteria considering that correlation gives a normalized output, whereas MI depends on the entropy of the distributions. 

\begin{figure*}[t]
\centering
\begin{subfigure}{4.2cm}	
	\centering
	\includegraphics*[trim=0 0 0 0,clip=true,width=4.2cm, height=1.8cm]{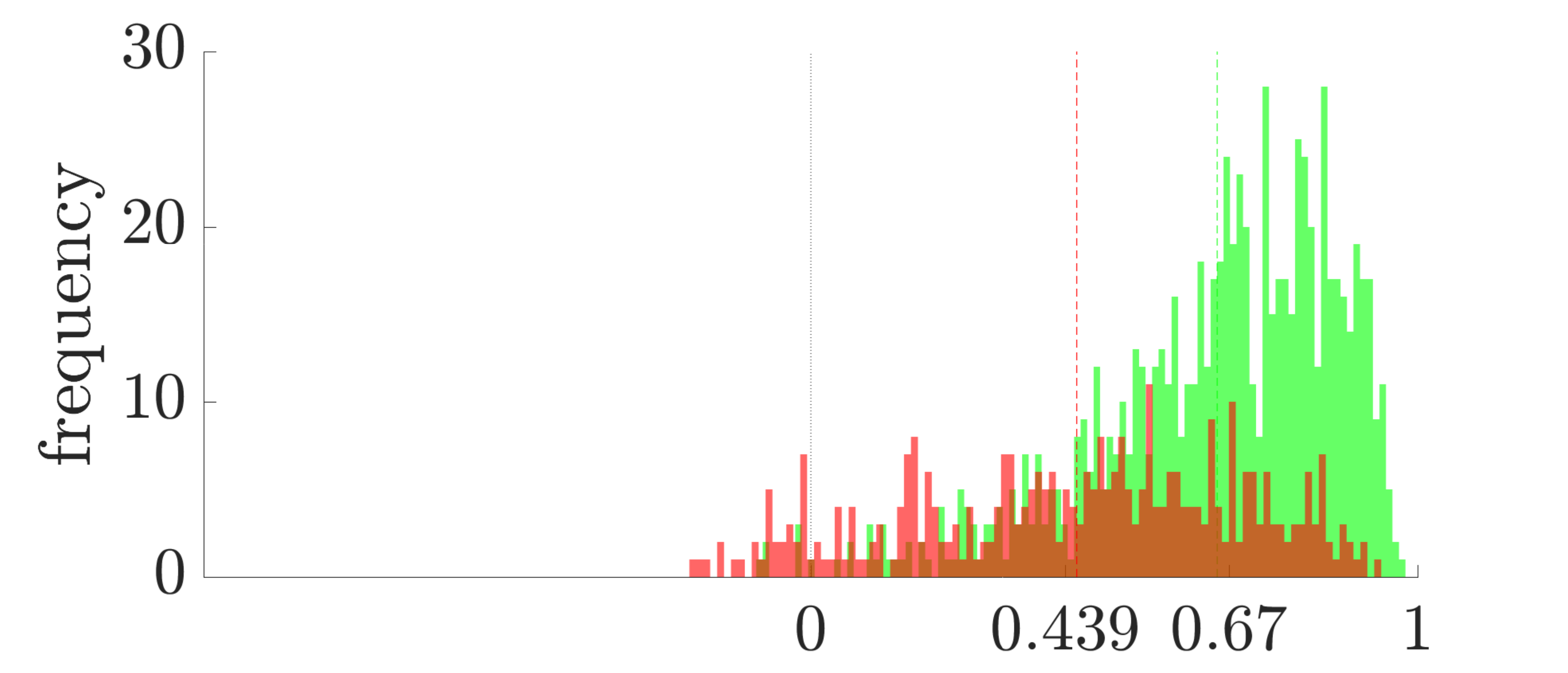}\par
	\caption{\(\rho\)$^{2D}$ histograms for true-green and false-red VMSs.}
\end{subfigure}
\begin{subfigure}{0.12cm}	
	\centering
	\includegraphics*[trim=0 0 0 0,clip=true,width=1cm, height=2.15cm]{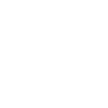}\par
\end{subfigure}
\begin{subfigure}{4.2cm}	
	\centering
	\includegraphics*[trim=0 0 0 0,clip=true,width=4.2cm, height=1.8cm]{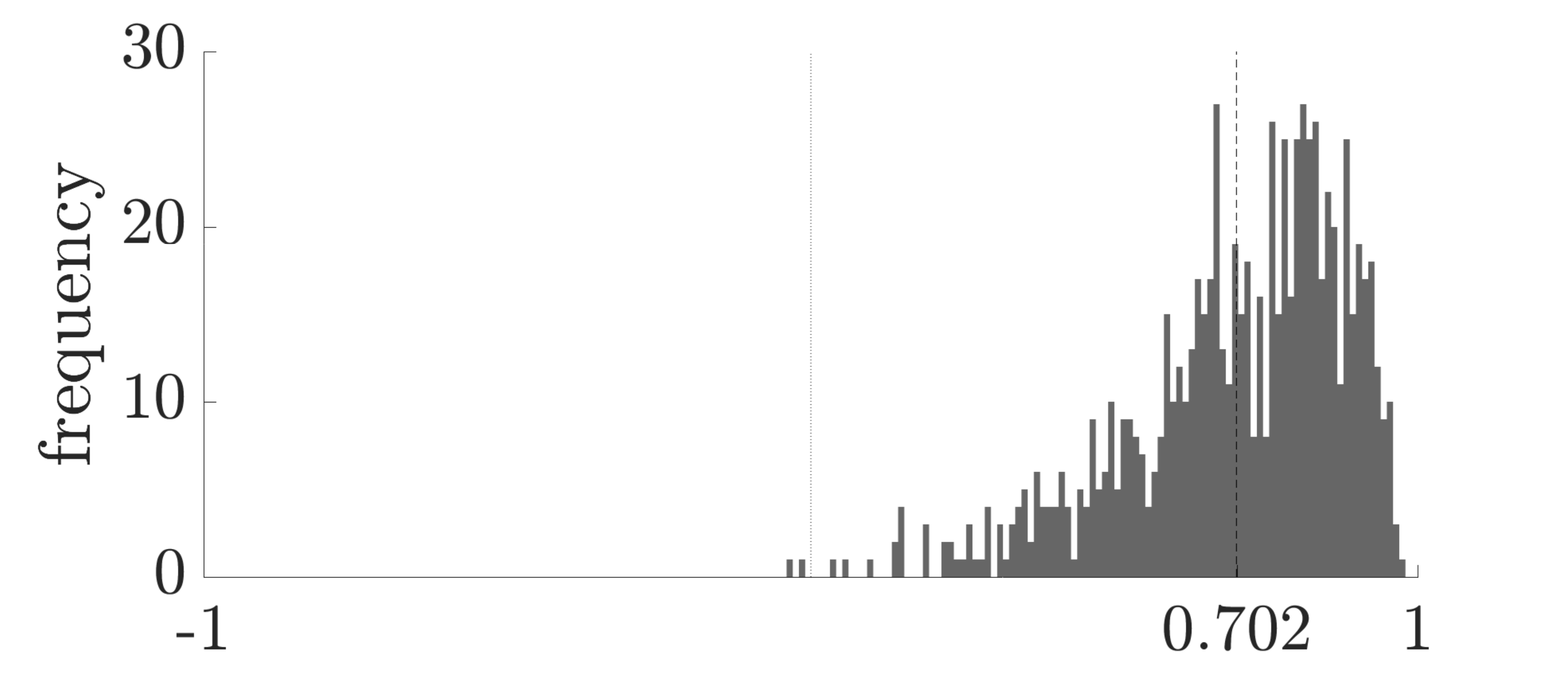}\par
	\caption{\(\rho\)$^{2D}$ histogram for the combined VMSs.}
\end{subfigure}
\begin{subfigure}{0.12cm}	
	\centering
	\includegraphics*[trim=0 0 0 0,clip=true,width=1cm, height=2.15cm]{figures/blank-eps-converted-to.pdf}\par
\end{subfigure}
\begin{subfigure}{4.2cm}	
	\centering
	\includegraphics*[trim=0 0 0 0,clip=true,width=4.2cm, height=1.8cm]{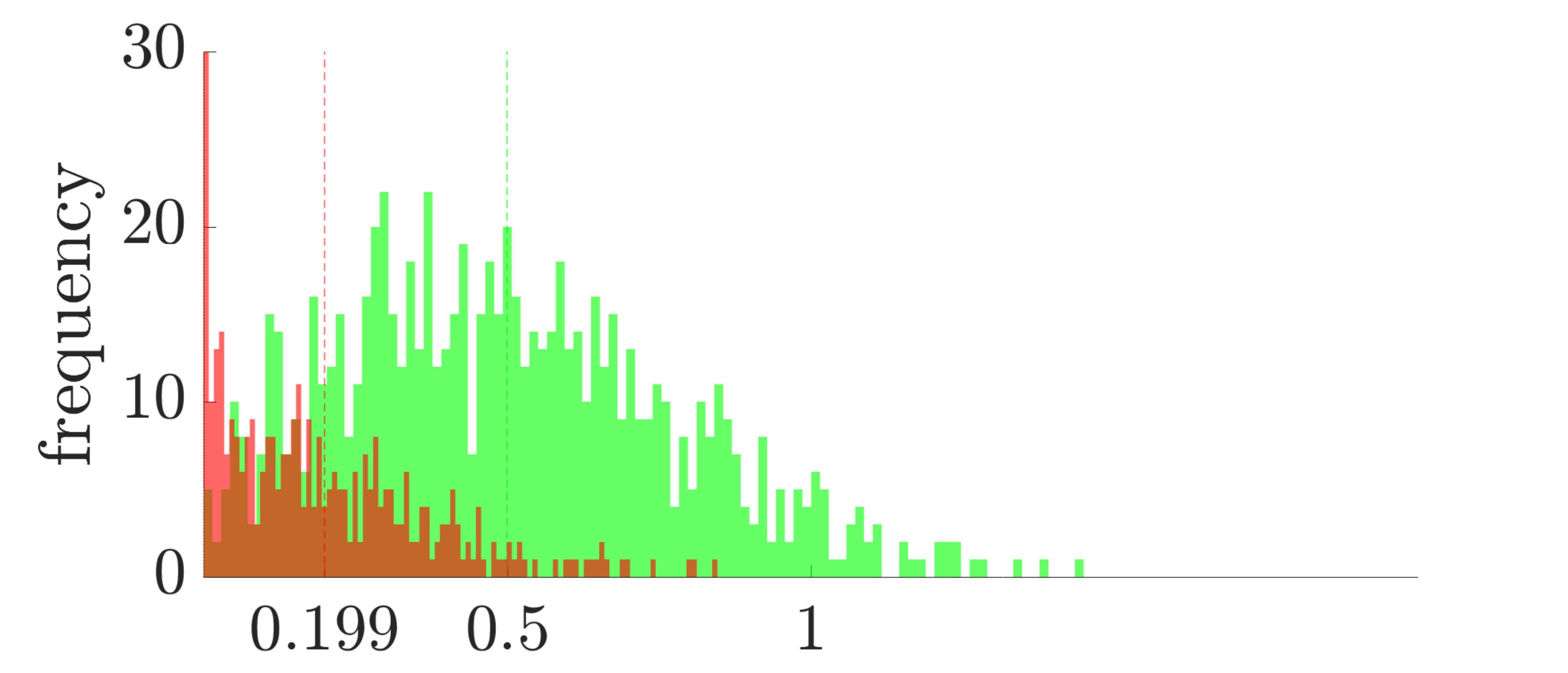}\par
	\caption{MI histograms for true-green and false-red VMSs.}
\end{subfigure}
\begin{subfigure}{0.12cm}	
	\centering
	\includegraphics*[trim=0 0 0 0,clip=true,width=1cm, height=2.15cm]{figures/blank-eps-converted-to.pdf}\par
\end{subfigure}
\begin{subfigure}{4.2cm}	
	\centering
	\includegraphics*[trim=0 -5 0 0,clip=true,width=4.2cm, height=1.8cm]{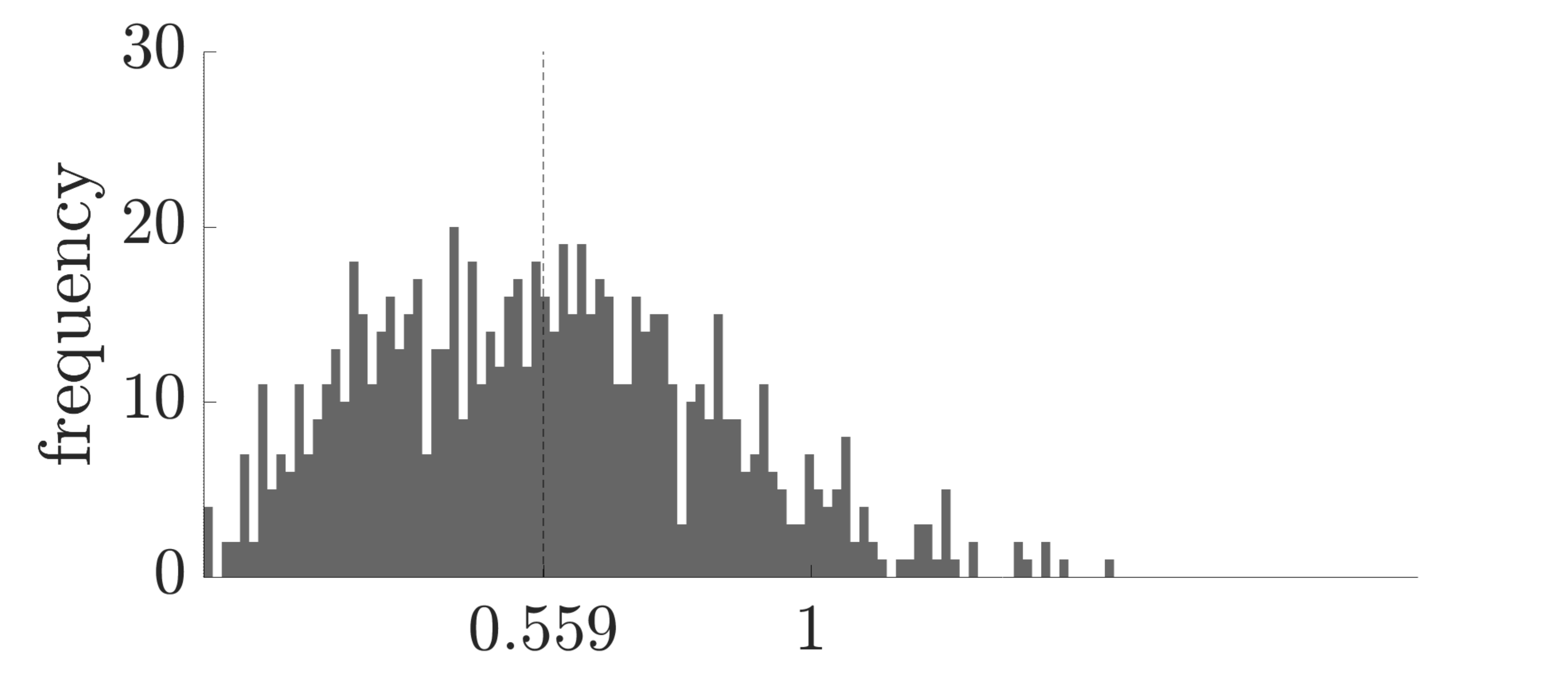}\par
	\caption{MI histogram for the combined VMSs.}
\end{subfigure}
\caption{Histograms of the average correlation (\(\rho\)$^{2D}$) and MI, between the VMS's corresponding to two equally sized groups of participants. The green histograms represent the correlation or MI for the true VMSs, while the red histograms correspond to those of the false VMSs.}
 \label{PartCorr}
\end{figure*}

\subsection{Analysis of VMS Consistency}
In order to examine how {strongly} a VMS is shared among different observers, one must first show that it is a consistent signal among different observers. For this purpose, participants are split into two randomly selected independent sets, equal in numbers. {For each image, two different VMS maps are obtained from each split set. The correlation and MI between the two different VMSs of each image are calculated separately for both true and false VMSs. This procedure is rerun for 25 different random splits and the average correlation and MI of 25 runs are calculated.} Histograms of the resulting average correlations and average MI values are shown in Figures \ref{PartCorr}.a-b and \ref{PartCorr}.c-d, respectively. 

\begin{figure*}[t]
 \centering
\begin{subfigure}{4.2cm}	
	\centering
	\includegraphics*[trim=0 0 0 0,clip=true,width=4.2cm, height=1.8cm]{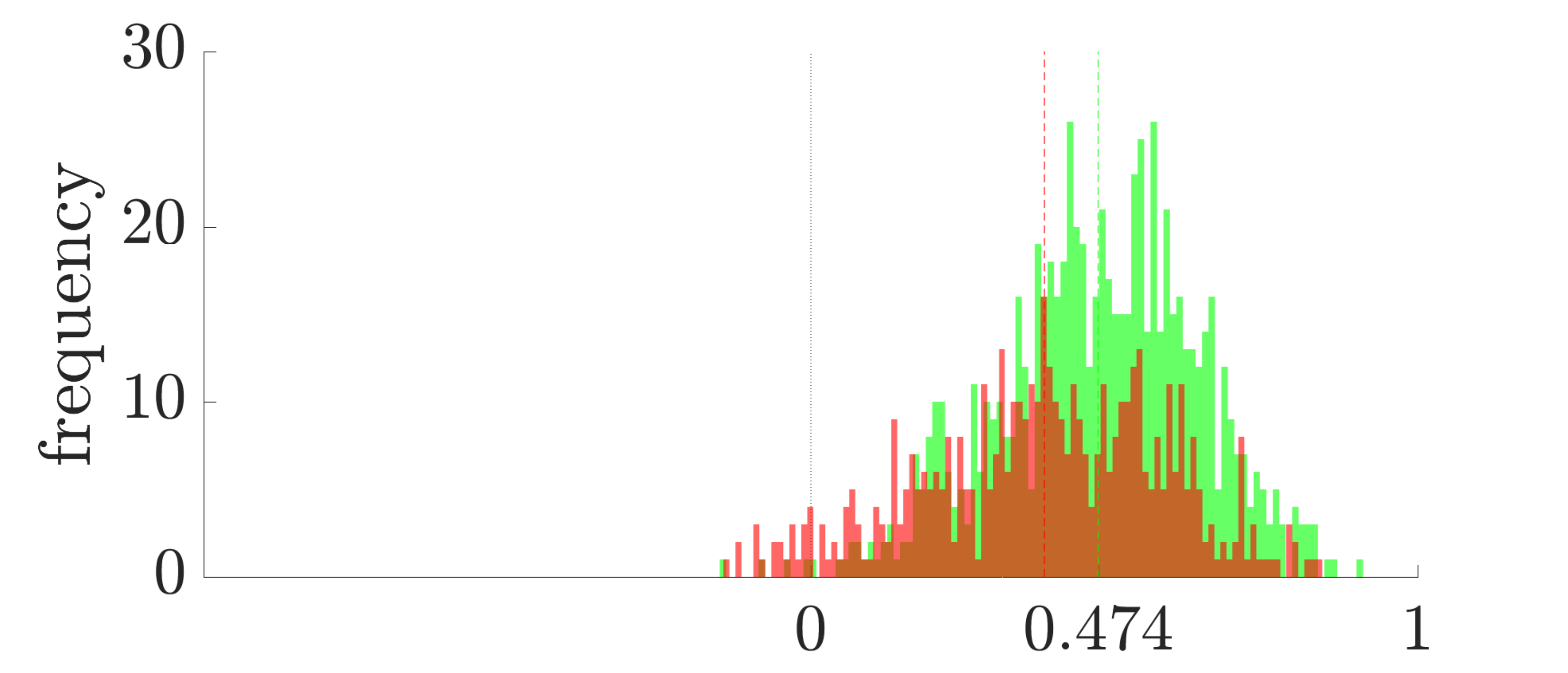}\par
	\caption{\(\rho\)$^{2D}$ histograms between the true-green and false-red VMSs and the eye fixations maps.}
\end{subfigure}
\begin{subfigure}{0.12cm}	
	\centering
	\includegraphics*[trim=0 0 0 0,clip=true,width=1cm, height=1.8cm]{figures/blank-eps-converted-to.pdf}\par
\end{subfigure}
\begin{subfigure}{4.2cm}	
	\centering
	\includegraphics*[trim=0 0 0 0,clip=true,width=4.2cm, height=1.8cm]{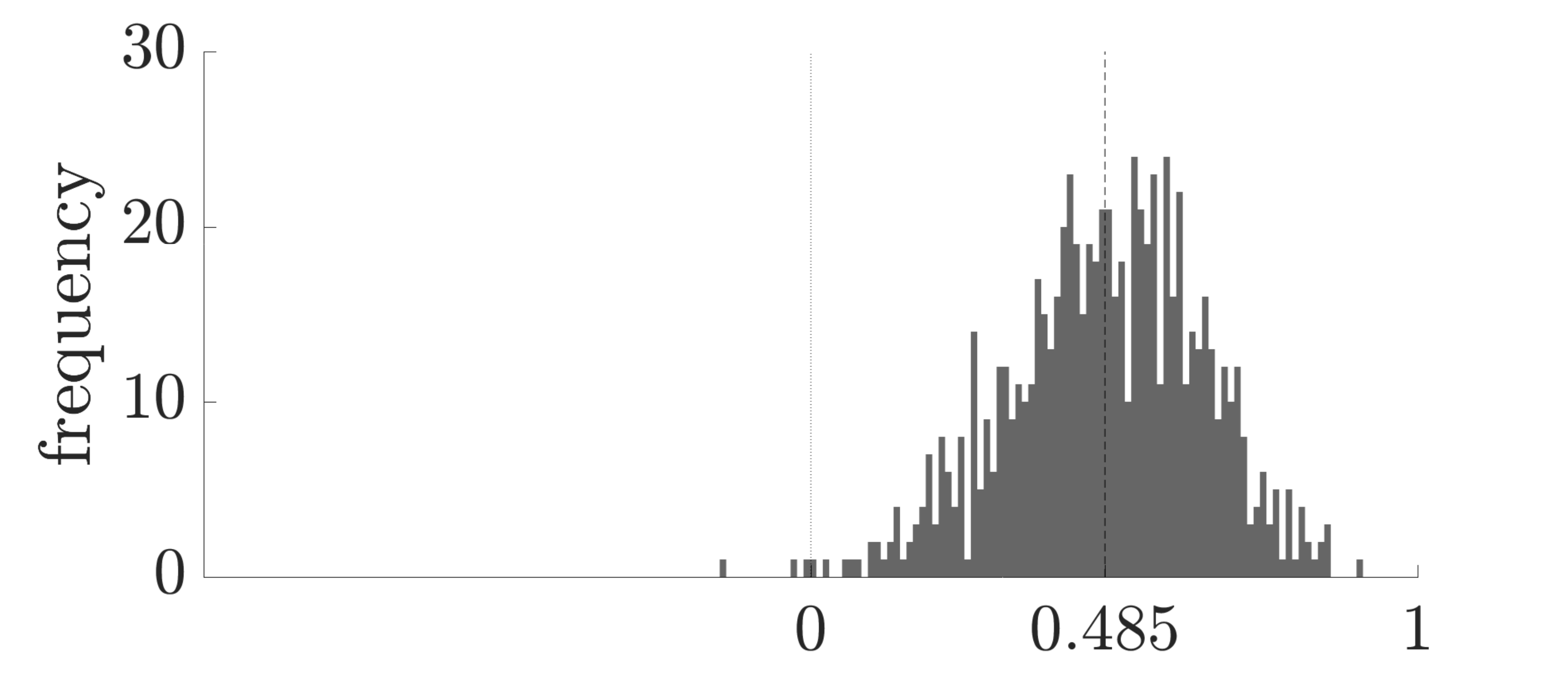}\par
	\caption{\(\rho\)$^{2D}$ histogram between the overall combined VMSs and the eye fixation maps.}
\end{subfigure}
\begin{subfigure}{0.12cm}	
	\centering
	\includegraphics*[trim=0 0 0 0,clip=true,width=1cm, height=1.8cm]{figures/blank-eps-converted-to.pdf}\par
\end{subfigure}
\begin{subfigure}{4.2cm}	
	\centering
	\includegraphics*[trim=0 0 0 -70,clip=true,width=4.2cm, height=1.8cm]{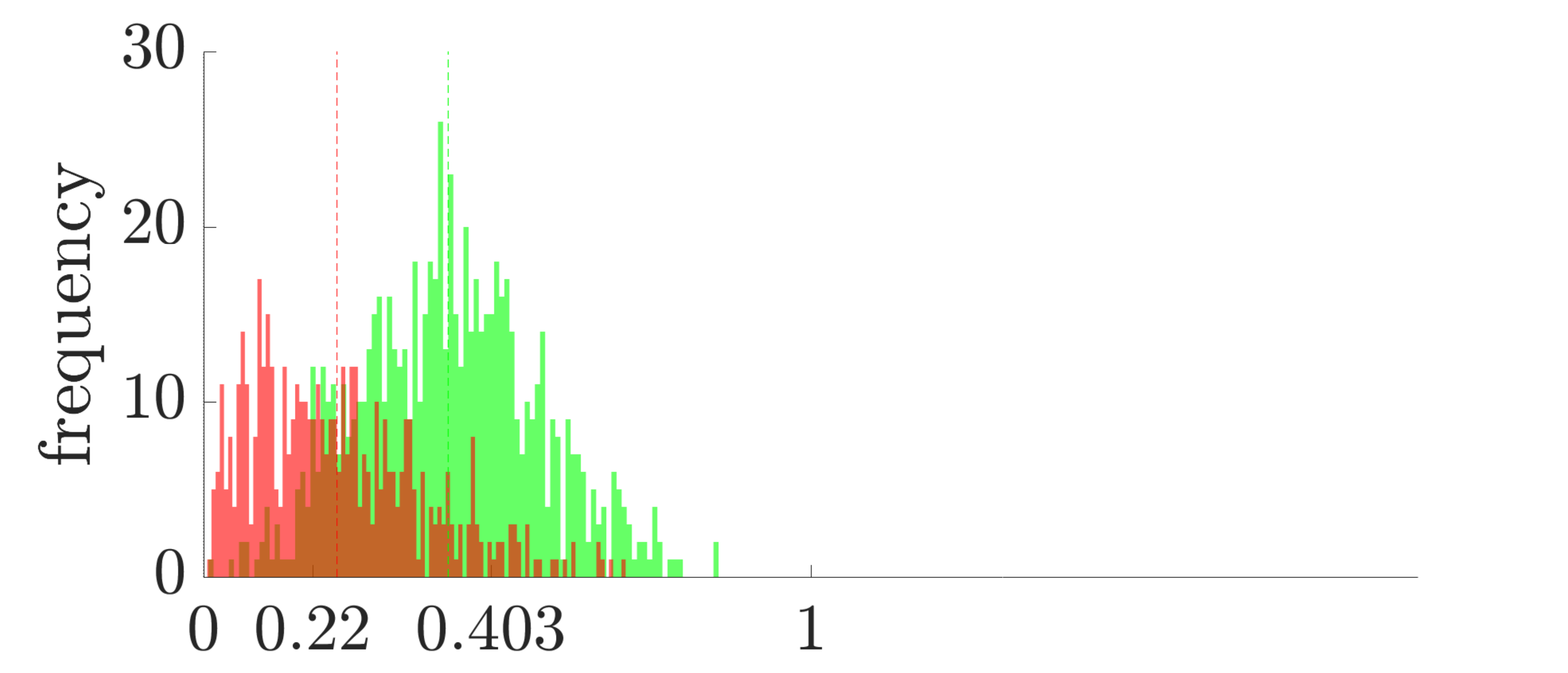}\par
	\caption{MI histograms between the true-green and false-red VMSâ€™s and the eye fixations maps.}
\end{subfigure}
\begin{subfigure}{0.12cm}	
	\centering
	\includegraphics*[trim=0 0 0 0,clip=true,width=4.2cm, height=1.8cm]{figures/blank-eps-converted-to.pdf}\par
\end{subfigure}
\begin{subfigure}{4.2cm}	
	\centering
	\includegraphics*[trim=0 0 0 -40,clip=true,width=4.2cm, height=1.8cm]{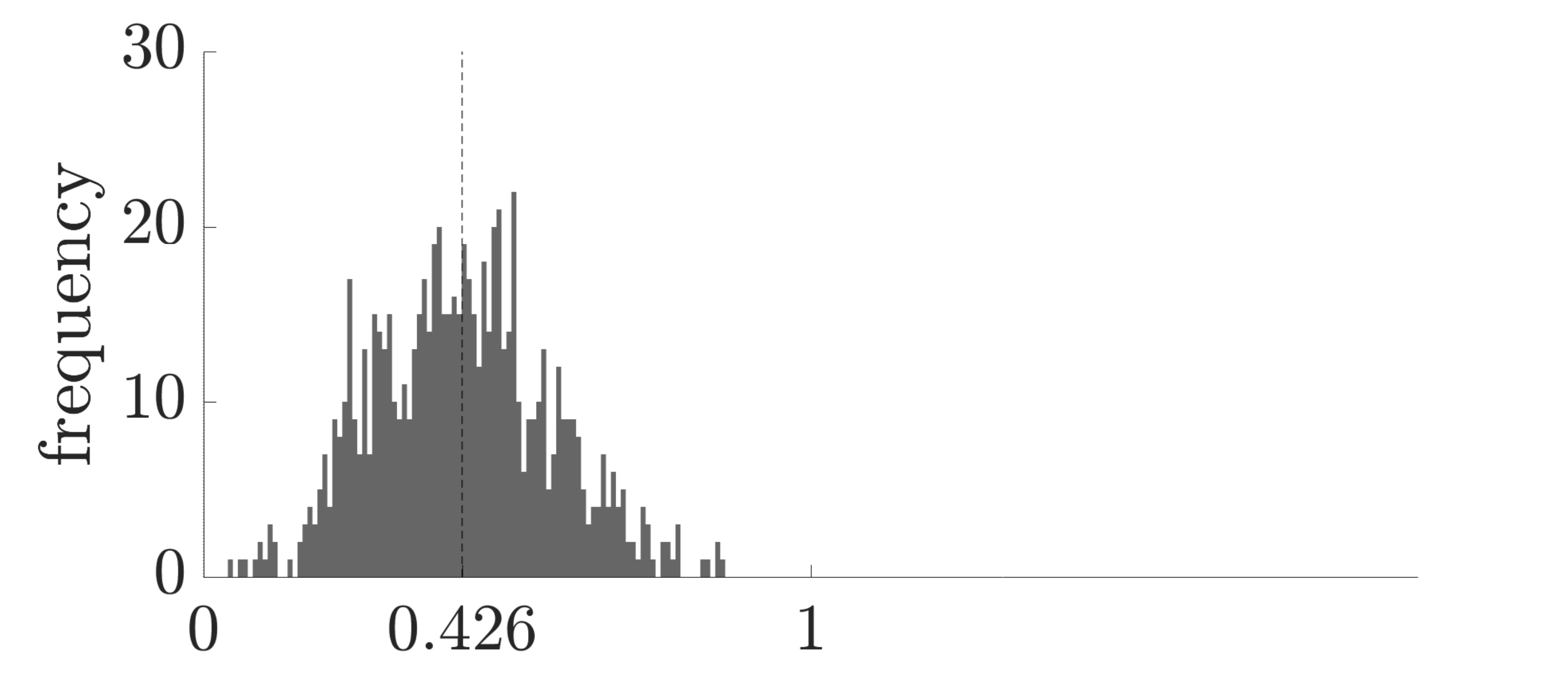}\par
	\caption{MI histogram between the overall combined VMSs and the eye fixation maps.}
\end{subfigure}
\caption{Correlation (\(\rho\)$^{2D}$) and MI histograms between the VMSs and eye fixations maps are depicted separately in green for the true VMS, in red for the false VMS and the combined VMSs in black. 
}
\label{InterEye}
\end{figure*}

\begin{figure*}[t]
 \centering
\begin{subfigure}{4.2cm}	
	\centering
	\includegraphics*[trim=0 0 0 0,clip=true,width=4.2cm, height=1.8cm]{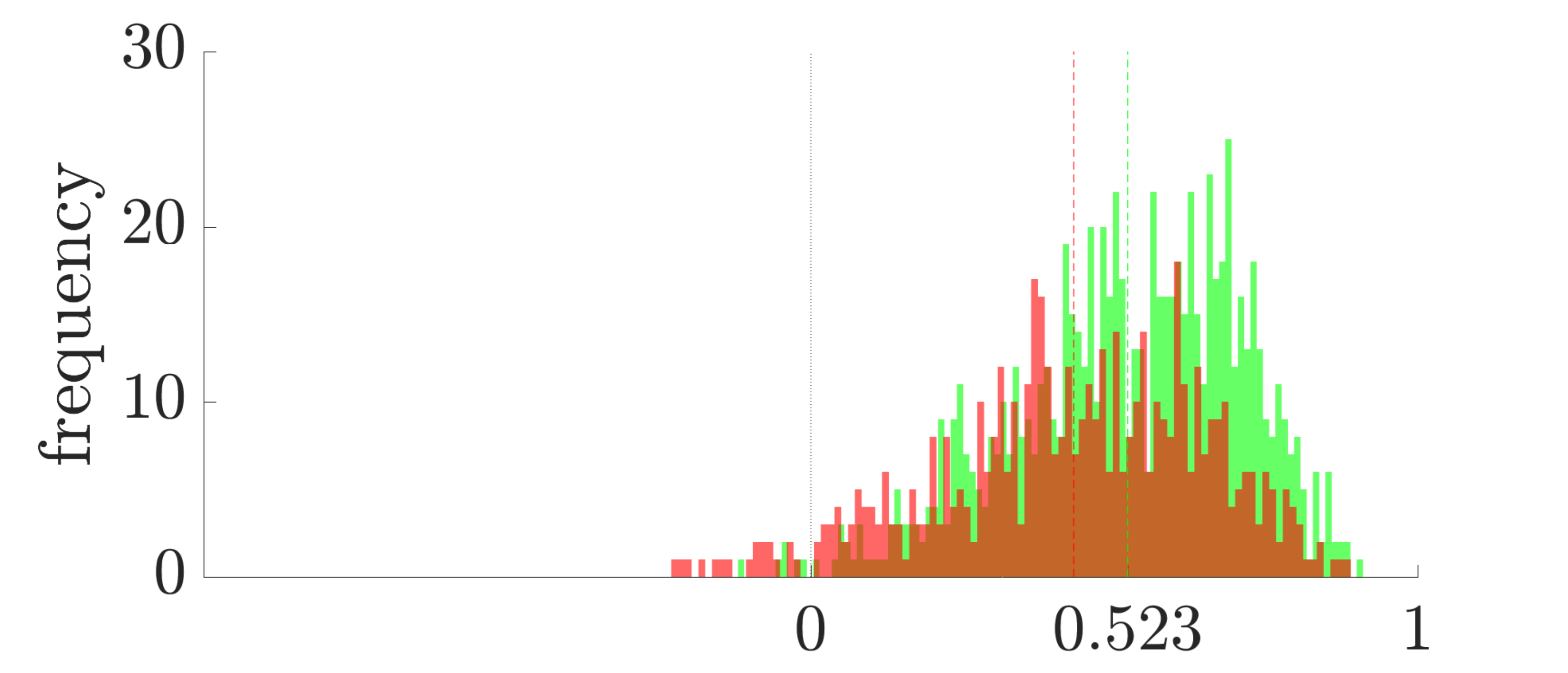}\par
	\caption{\(\rho\)$^{2D}$ histograms between the true-green and false-red VMSs and the saliency maps.}
\end{subfigure}
\begin{subfigure}{0.12cm}	
	\centering
	\includegraphics*[trim=0 0 0 0,clip=true,width=1cm, height=1.8cm]{figures/blank-eps-converted-to.pdf}\par
\end{subfigure}
\begin{subfigure}{4.2cm}	
	\centering
	\includegraphics*[trim=0 0 0 0,clip=true,width=4.2cm, height=1.8cm]{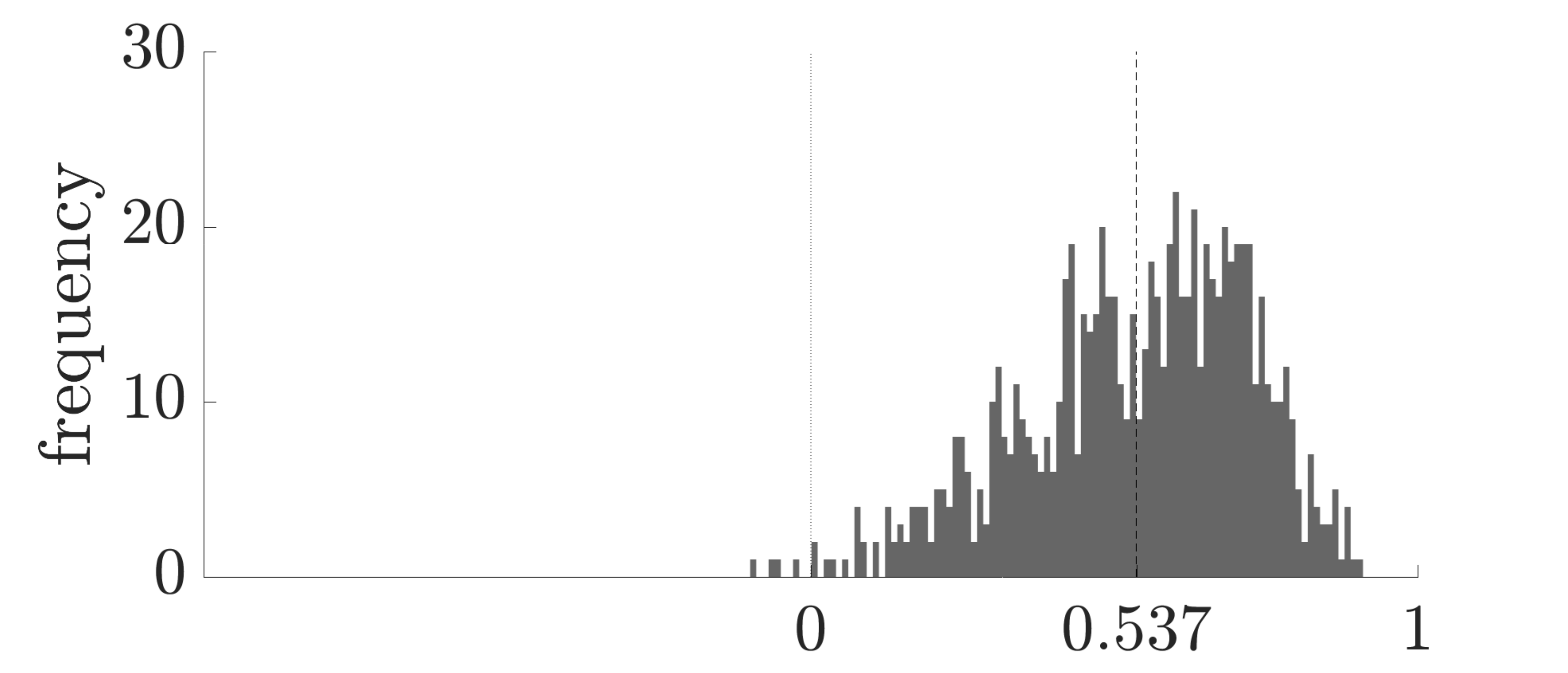}\par
	\caption{\(\rho\)$^{2D}$ histogram between the overall combined VMSs and the saliency maps.}
\end{subfigure}
\begin{subfigure}{0.12cm}	
	\centering
	\includegraphics*[trim=0 0 0 0,clip=true,width=1cm, height=1.8cm]{figures/blank-eps-converted-to.pdf}\par
\end{subfigure}
\begin{subfigure}{4.2cm}	
	\centering
	\includegraphics*[trim=0 0 0 0,clip=true,width=4.2cm, height=1.8cm]{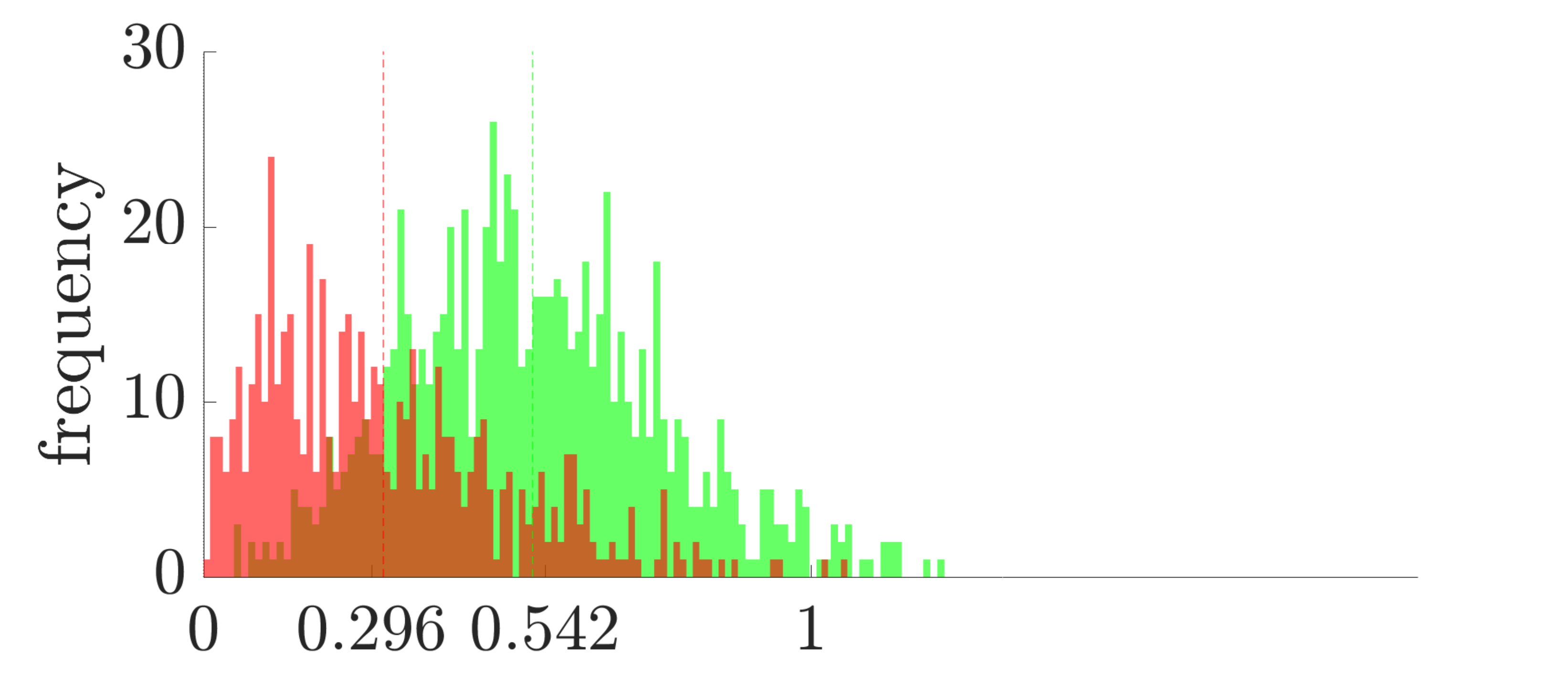}\par
		\caption{MI histograms between the true-green and false-red VMSs and the saliency maps.}
\end{subfigure}
\begin{subfigure}{0.12cm}	
	\centering
	\includegraphics*[trim=0 0 0 0,clip=true,width=1cm, height=1.8cm]{figures/blank-eps-converted-to.pdf}\par
\end{subfigure}
\begin{subfigure}{4.2cm}	
	\centering
	\includegraphics*[trim=0 0 0 0,clip=true,width=4.2cm, height=1.8cm]{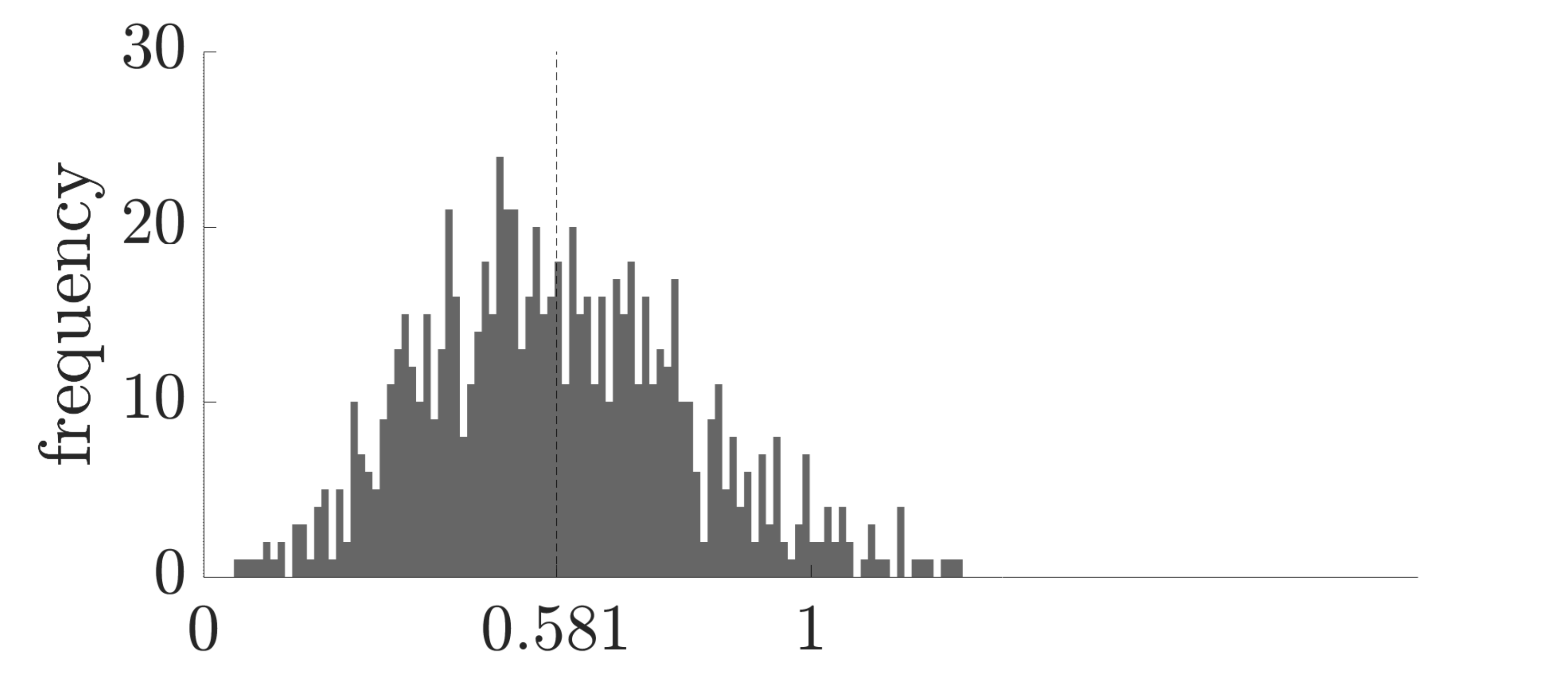}\par
\caption{MI histogram between the overall combined VMSs and the saliency maps.}
\end{subfigure}
\caption{ Correlation (\(\rho\)$^{2D}$) and MI histograms between VMS and saliency maps are depicted separately for true VMS , shown in green, false VMS, shown in red, and combined VMSs, shown in black. 
}
\label{InterSel}
\end{figure*}

The correlation histogram for true VMSs (Figure \ref{PartCorr}.a green) with a mean (\(\mu\)) of 0.67 and standard deviation (\(\sigma\))\footnote{Please note that this \(\mu\) and \(\sigma\) are the mean and standard deviation of the histograms for Figure \ref{PartCorr}.a. In this and the following two subsections, the symbols \(\mu\) and \(\sigma\) are always used to indicate the means and standard deviations of the "histograms" in figures \ref{PartCorr},  \ref{InterEye} and  \ref{InterSel}.} of $\pm$0.202, shows that the true VMS is highly consistent among observers. On the other hand, the correlation histogram for false VMSs  (Figure \ref{PartCorr}.a red) has a lower histogram mean of 0.439 (and a standard deviation of $\pm$0.272) and has negative values for some images. The results for the average MI histograms are quite similar to those of the correlations as seen in Figure \ref{PartCorr}.c. MI histogram for true VMSs (Figure \ref{PartCorr}.c green) shows higher dependency compared to MI histogram for false VMSs  (Figure \ref{PartCorr}.c red). 

In order to understand if the red and green histograms in Figures \ref{PartCorr}.a and \ref{PartCorr}.c are significantly different from each other, we apply a bootstrapping test on the difference between these two histograms. For this purpose, we calculate the sample-based difference between the two histograms for a randomly selected subset of images and check whether the difference values span zero value within a 95\% confidence interval. We repeat this test for 10,000 times and if, for any test, 95\% confidence interval of the difference values span 0, we conclude that the two histograms are not significantly different. We also use this bootstrapping technique in the statistical analyses provided in the following subsections.

We observe a {significant} difference (p$<$0.0001) between the distribution of correlations for true and false VMSs, suggesting that the memorability of images is based on different types of visual schemas for correctly and falsely remembered images. True VMSs are more consistent across observers indicating that they are based on widely shared knowledge and experience when compared to false VMSs. {Consequently it can be hypothesized that observers use more established schemas or so called prototypical schemas, reliant on semantic knowledge when encoding the correctly recalled images.} Whereas, when observers falsely recognized images they rather relied on visual schemas, rather derived from individual episodic experience.

Figure \ref{PartCorr}.b and \ref{PartCorr}.d show the correlation and MI histograms, respectively for combined VMSs. For this purpose, all VMS selections are combined regardless of being true or false. The mean of the combined correlation histogram is 0.7, much higher than the true or false VMS histograms\footnote{When calculating the distances among false VMSs, the empty selections, {\em i.e.} the images with no false selections, are omitted because it is not possible to calculate the correlation or MI for them. However when calculating the combined VMSs, empty false VMSs are included, since they are always a part of a group of non-empty true and false VMSs. That's the reason why the combined VMS consistency is higher than the sum of false and true VMSs.} considered individually. We use this correlation value as a benchmark when we compare the similarity of visual saliency and eye fixations to VMS in the following subsections.

\subsection{Analysis of VMS and Eye Fixation Relationship}
In this section we examine the relationship between VMS and observers' eye fixation that stand as proxy to overt attention. For this purpose, we calculate the distance between VMS maps of an image and the eye fixations for the same images but from a different group of observers. For a majority of 745 images, out of the total of 800 from the VISCHEMA image set, we have the corresponding eye fixation maps. For these images we calculate the correlation (\(\rho\)$^{2D}$) and MI between the eye fixation maps and VMSs. Figures \ref{InterEye}.a-b and \ref{InterEye}.c-d show the correlation and MI histograms respectively and separately for true (green), false (red) and combined (black) VMSs.

The correlation histograms between the VMS and eye fixations for true VMSs (Figure \ref{InterEye}.a), is $\mu$=0.474 and $\sigma$=$\pm$0.166 and for false VMSs is $\mu$=0.385 and $\sigma$=$\pm$0.198. From this plot it is evident that the average of the correlation histograms for both the true and false VMS are quite similar and both have values of less than 0.5, which are lower when compared to VMS self-distance consistency of 0.7. This indicates that neither type of VMS is highly correlated with eye fixations. {We see the same pattern for MI histograms with means relatively closer to 0 when compared to VMS self-distance consistency, as shown in Figure }\ref{InterEye}.c. Figures \ref{InterEye}.b and \ref{InterEye}.d display the same histograms for combined VMSs and support the same conclusions. Bootstrapping tests confirm that VMS and eye fixation location distributions differ significantly from each other (p$<$0.0001). These results indicate that {VMS can not be fully explained by overt attention.}

\subsection{Analysis of VMS and Saliency Relationship}
In the following we examine the relationship between VMS and visual saliency, as defined by graph-based visual saliency algorithm (GBVS) \cite{Harel2006}, for the 800 VISCHEMA images. GBVS is a bottom-up visual saliency model, which models computationally the visual saliency in images. The algorithm creates Markov chains over image maps and treats the equilibrium distribution over map locations as saliency values. For all 800 images from the VISCHEMA Image set, we construct $100 \times 100$ resolution graph-based visual saliency (GBVS) maps using the algorithm proposed in Harel et al. \cite{Harel2006}. After constructing the GBVS maps for the VISCHEMA Image set, the correlation (\(\rho\)$^{2D}$) and MI between the saliency, on one hand, and the true VMSs, the false VMSs and  the combined VMSs of each image, on the other hand, are calculated and shown in Figures \ref{InterSel}.a-b and \ref{InterSel}.c-d, respectively.

It can be observed from the histograms from Figure 8 that there is a non-significant relationship between the VMSs and the visual saliency. Bootstrapping tests show that{, compared to VMS consistency of 0.702,} there is no strong correlation between saliency and the VMSs with an average correlation distance of 0.581 for the combined VMSs versus the visual saliency. This is far less than the average correlation self-distance of 0.7 for the combined VMSs, as shown in Figure \ref{PartCorr}. {Thus, we conclude that visual saliency does not fully account for the proposed visual memory schema concept.}\footnote{{The reader should note that the tests are carried out for a single type of visual saliency algorithm, namely GBVS, and results may vary if a different algorithm is used for computationally modelling the saliency in images.}}

\section{Image Memorability Tests}
Previous studies \cite{Isola2011a} have shown that computer vision features can be used to predict image memorability with rank correlations of up to 0.5. Larger scale experiments using convolutional neural networks \cite{Khosla2015} show that such results can be further improved. {However, despite the improvements in the memorability rates achieved in such research studies, they do not fully explain what makes an image memorable. In this section we focus our analysis on the role played by the proposed VMS concept in image memorability. More specifically we assess how computer vision features are more effective when they are spatially pooled within a VMS. Moreover, we analyse the prediction results produced for each scene category, separately.}

\subsection{Test Setup}
In this section we assess the contribution of visual memory schemas when using machine learning algorithms to predict image memorability. Similarly to previous studies \cite{Isola2011a}, we use Spearman's rank correlation and the average empirical memorability scores for the top 20, top 100, bottom 100 and bottom 20 images, selected by the machine learning algorithms for their memorability. The performance is evaluated over 25 random splits of the VISCHEMA image dataset, each split containing an equal number of 400 training and 400 testing images, in order to make these results consistent with those provided by Isola et al. in \cite{Isola2011a}. These training and testing splits were scored by different halves of the participants. The results indicate consistency among observers with a rank correlation of 0.5, when using the threshold of 40, which was adopted as explained in section Section III.C. The effectiveness of the prediction models are assessed by comparing the rank correlation results to this score.

In order to estimate the memorability scores, a support vector regression (SVR) machine is trained using LibSVM \cite{LibSVM2011}. Various well known computer vision features, such as pixel histograms, the spatial envelope (GIST) \cite{Oliva2001}, the scale invariant feature transform (SIFT) \cite{Lowe2004} and histograms of gradients (HoG) \cite{Dalal2005} are used to create feature vectors for the images. Similarly to the study from Isola et al. \cite{Isola2011a}, we use RBF kernels for modelling the GIST features, histogram intersection kernels for the pixel histograms, SIFT and HoG, and a kernel product for the combination of these features. The code used for the calculation of all these features is available from our website \footnoteref{website}.

\begin{table}[t]
     \centering
\caption{The performance of computer vision features on predicting image memorability and human consistency.}
\label{ResultTable1}
     \begin{tabular}{|c|c|c|c|c|c|c|c|c|} \hline
 & Humans  & Pixels & GIST & SIFT & HoG  & Comb. \\ \hline
Top 20 & 67.86 & 47.48 & 53.29 & 54.35 & 54.07 &  56.10  \\ \hline
Top 100 & 59.12 & 46.90  & 49.41 & 50.49 & 50.07 & 51.05  \\ \hline
Bottom 100 & 33.14 &44.17  & 39.76 & 38.77 & 37.23  & 37.90  \\ \hline
Bottom 20 & 28.10 & 42.24 & 34.27 & 31.84 & 30.67 & 32.15  \\ \hline
 \(\rho\) & 0.50 & 0.044 &  0.19 & 0.20 & 0.24 &  0.24 \\ \hline
      \end{tabular}
    \end{table}

\begin{table}[t]
     \centering
\caption{The performance of computer vision features on predicting image memorability when spatially pooled and weighted with  \textbf{saliency maps}.} 
\label{ResultTable3}
     \begin{tabular}{|c|c|c|c|c|c|} \hline
       & {Sal.} \& & {Sal.} \& & {Sal.} \& & {Sal.} \& & {Sal.} \& \\
       & Pixels & GIST & SIFT & HoG & Combined \\ \hline
Top 20 & 46.70 & 50.29 & 53.12 & 51.71 & 51.86 \\ \hline
Top 100 & 46.11 & 47.89 & 50.09 & 48.88 & 50.32  \\ \hline
Bottom 100 & 43.48 & 40.30 & 38.29 & 36.88 & 37.51\\ \hline
Bottom 20 & 41.84 & 37.04 & 32.17 & 29.72 &  30.15\\ \hline
 \(\rho\) & 0.052  & 0.15 & 0.22 & 0.21 & \underline{\textbf{0.25}}\\ \hline
      \end{tabular}
    \end{table}

\begin{table}[t]
     \centering
\caption{The performance of computer vision features on predicting image memorability when spatially pooled and weighted with \textbf{eye-fixation maps}.} 
\label{ResultTable4}
     \begin{tabular}{|c|c|c|c|c|c|} \hline
       & {Eye Fix.} \& & {Eye} \& & {Eye} \& & {Eye} \& & {Eye} \& \\
       & Pixels & GIST & SIFT & HoG & Combined \\ \hline
Top 20 & 46.80 & 51.82 & 51.48 & 54.44 & 53.38 \\ \hline
Top 100 & 46.43 & 48.78 & 49.84 & 51.58 & 50.93  \\ \hline
Bottom 100 & 43.07 & 42.05 & 38.67 & 37.05 & 37.36\\ \hline
Bottom 20 & 41.36 & 43.37 & 32.16 & 30.51 &  31.25\\ \hline
 \(\rho\) & 0.054  & 0.13 & 0.21 & 0.29 & \underline{\textbf{0.29}}\\ \hline
      \end{tabular}
    \end{table}

\begin{table}[t]
     \centering
\caption{The performance of computer vision features on predicting image memorability when spatially pooled and weighted with \textbf{VMS selections}.} 
\label{ResultTable2}
     \begin{tabular}{|c|c|c|c|c|c|} \hline
       & {VMS} \& & {VMS} \& & {VMS} \& & {VMS} \& & {VMS} \& \\
       & Pixels & GIST & SIFT & HoG & Combined \\ \hline
Top 20 & 48.64 & 47.94 & 51.39 & 59.32 & 61.25 \\ \hline
Top 100 & 46.95 & 47.33 & 49.85 & 53.45 & 55.42  \\ \hline
Bottom 100 & 42.56 & 41.78 & 38.70 & 35.26 & 33.48\\ \hline
Bottom 20 & 41.06 & 39.42 & 32.67 & 27.40 &  24.87\\ \hline
 \(\rho\) & 0.085  & 0.10 & 0.21 & 0.34 & \underline{\textbf{0.41}}\\ \hline
      \end{tabular}
    \end{table}

\subsection{Using computer vision features from entire images as inputs to Machine Learning Algorithms}
VISCHEMA image set is a subset of the FIGRIM and SUN image sets, as mentioned in Section 3, while certain images known to be highly memorable, are deliberately excluded. For this reason, the average memorability scores obtained from human observers for the VISCHEMA image set are lower then those obtained in other memory experiments. Moreover, the human consistency in our experiment is also lower, which is expected when images are hard to remember. Table \ref{ResultTable1} shows the prediction results on the VISCHEMA image set using computer vision features calculated from entire images as in the study from \cite{Isola2011a}. The rank correlations calculated previously on a different image set reported in \cite{Isola2011a}, are \(\rho\)$_{Pixels}$=\emph{0.22}, \(\rho\)$_{GIST}$=\emph{0.38}, \(\rho\)$_{SIFT}$=\emph{0.41}, \(\rho\)$_{HoG}$=\emph{0.43}, \(\rho\)$_{Comb}$=\emph{0.46}. When we compare these results to those from Table  \ref{ResultTable1}, we can observe that the prediction results are much lower for the VISCHEMA image set. When the stimuli set becomes challenging, in other words, when the easily memorable images are left out, the results obtained from the human observers fall by 10\%, representing a significant drop in memorability. Thus, it is expected that simple computer vision features, which lack the semantic and syntactic information description of the image, would provide a low performance for predicting image memorability.

Tables \ref{ResultTable3} and \ref{ResultTable4} provide the results when considering the computer vision features pooled with saliency and eye-fixation maps, respectively. It can be observed from these tables, that pooling with saliency maps, generated by the GBVS algorithm, does not increase the prediction success of the computer vision features, whereas pooling with eye-fixation maps would show an increase of only 5\% in performance results.
 
\subsection{The Significance of VMS for Image Memorability in Machine Learning Tests}
Here we use the VMS selections for spatially pooling the computer vision features. Similarly to the procedure described in the previous section, after creating a histogram for each of these features, its frequency for each bin is weighted by the value of the average VMS selections falling into that bin. In this way the computer vision features are spatially pooled and their effect on predicting image memorability is weighted by the VMS selections. The results from Table \ref{ResultTable2}, indicate that, when using spatially pooled and weighted by the VMS selection values, computer vision features' prediction performance is considerably increased, when compared to the results provided by pooling the features from entire images. By using a kernel product for representing GIST, SIFT and HoG features, the SVR can predict image memorability with a rank correlation of 0.41, which is close to the rank correlation of 0.5, obtained for the human observers. This significant result shows that the overall VMS of an image represents a spatially refined visual signal that carries the information related to the memorability of an image.

\begin{figure}[t]
 \centering
	\includegraphics*[trim=10 0 -10 0,clip=false,width=8cm, height=4cm]{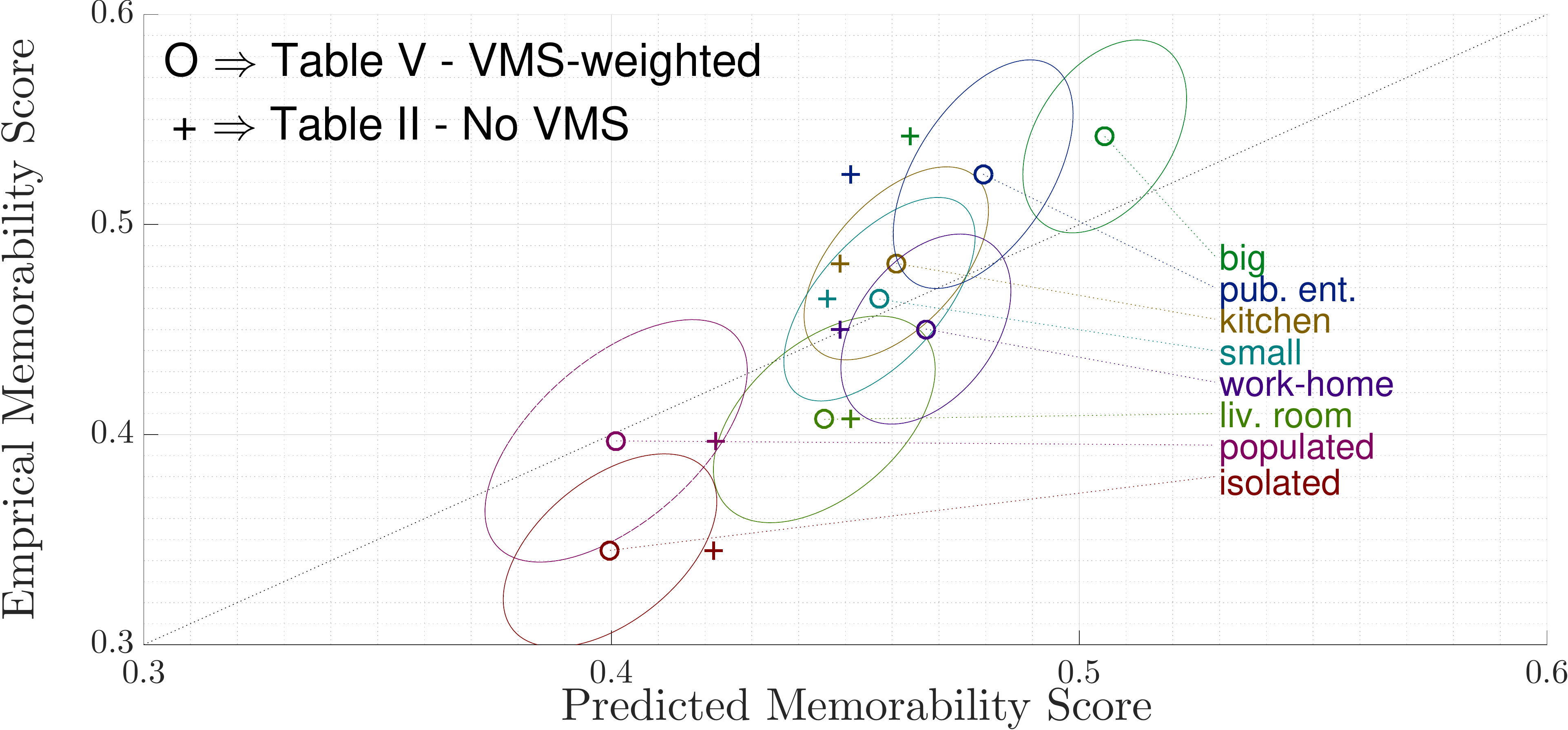}\par
	\caption{
Plotting predicted against empirically obtained memorability scores for each image category. The circles are the average predictions results for VMS-weighted features, as shown in Table \ref{ResultTable2}), whereas the plus signs are the average prediction results when No VMS is used, as shown in Table \ref{ResultTable1}). The ellipses, drawn around the small central circles, indicate the error spreads of the VMS-weighted features with widths corresponding to three standard deviations in the direction of each eigen-vector.}
 \label{CatResMemo}
\end{figure}

Next we focus our analysis on image category-based results. In order to understand how the VMS contributes to predicting image memorability, we compare the results of the proposed VMS pooled features with the approach when using entire images, for each image category. The predicted memorability scores, when using VMSs, for each image category, are compared against the empirical memorability and the results are shown in Figure \ref{CatResMemo}.  In this figure, the small circles indicate the average prediction results, obtained by using the VMS-weighted combination of GIST, SIFT and HoG features, as reported in the last column of Table \ref{ResultTable2}, for each category separately. These results are referred as "VMS $\&$ Combined". Similarly, the plus signs indicate the average prediction results when the combinations of the computer vision features are used without the VMS weighting, as reported in the last column of Table \ref{ResultTable1}, again for each category separately. These results are referred as "No VMS". The closer a circle or a plus sign is to the x=y diagonal line, the more successful is the average prediction for that image category. As it can be observed in Figure \ref{CatResMemo}, the circles are closer to the diagonal for almost all image categories. The ellipses, which are drawn around the circles as their centres, indicate the error spread for the VMS-weighted results. The widths of these ellipses represent three standard error deviations in the direction of each eigen-vector. By looking at the circles and plus signs, it is clear that the VMS-weighted features improve considerably the memorability predictions for all image categories with the exception of the \emph{work-home} category.

\begin{figure*}[!ht]
 \centering
	\includegraphics*[trim=5 0 -5 0,clip=false,width=17cm, height=6cm]{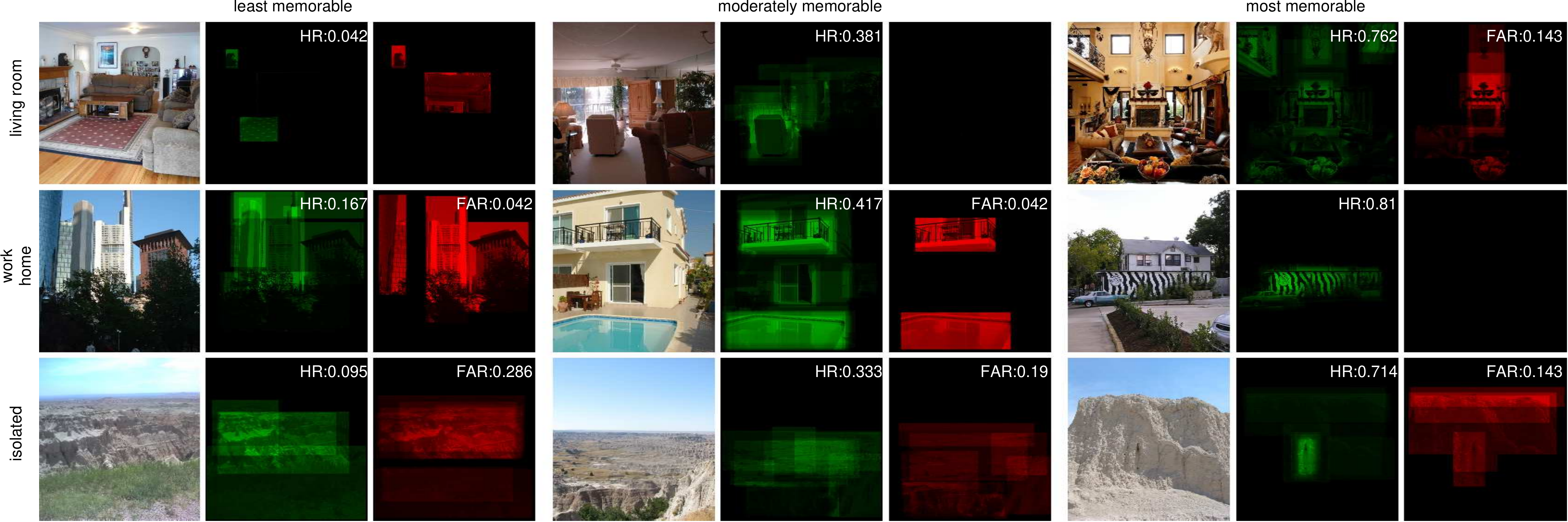}\par
	\caption{The least memorable, moderately memorable and the most memorable image, are shown together with true and false VMSs, where HR and FAR scores are indicated as well.}
 \label{CatRes}
\end{figure*}

In Figure \ref{CatRes} we provide examples of the least, the most, and the moderately memorable images from three image categories, together with their HR and FAR values and their true and false VMSs. Similarly to previous findings, such as those from \cite{Isola2011a} and \cite{Bylinskii2015}, it can be observed that while images with plain backgrounds and no objects are easily forgotten, images with specific, easily identifiable objects or with differentiating visual contexts are better remembered. For example, the least memorable image from the big image category, as shown in Figure \ref{CatRes}, completely lacks any objects. At the same time the most memorable image from the {\em work-home} image category has a very distinct wall colour. Both of these images are good examples of what may be the characteristic of either non-memorable or memorable images.

We can observe from Figure \ref{CatRes}, that images that have a stereotypical group of objects or a distinct organization of elements in the image that allows for the organization into visual schemas not only are better remembered but elicit also more false alarms. On the other hand a small difference between  the "No VMS" results as opposed to the "VMS $\&$ Combined" results for the \emph{living room} category images depicted in Figure \ref{CatResMemo}, indicates that VMS pooling does not contribute much to predicting memorability for this specific image category. This is because the visual schemas for the aforementioned image category are too general. For example every living room is expected to have a sofa, coffee table, artwork rug and these are usually colourful and in most cases located in the center of the room. The same logic can explain the results obtained for the \emph{work-home} category. In this category although images of houses or skyscrapers create distinctive image features, what people remember is actually the organization of visual schemas in the image and less so the features. The memorability of this image category is lower because its organization of VMS is not distinct at all. Another good example can be found within the \emph{isolated} scenes, which usually carry simple visual schema organizations like the \emph{work-home} scenes. These plain and relatively featureless images have the lowest prediction scores when considering only computer vision features, because, unlike the \emph{work-home} category, they lack the variation in feature diversity. However, computer vision features perform much better, when adding VMSs, even when there is a very simple but distinguishing visual schema organization that differentiates the image, such as "\emph{the white steep rock with a strange hole in it, under a blue sky}" and we observe that this is the most memorable image in the isolated category in Figure \ref{CatRes}.

{These results show that computer vision based memorability prediction algorithms can be improved by taking into account the visual schemas.} On the other hand, the organization of visual schemas within the image represent the defining information that makes an image memorable.

\section{Reconstructing the VMS Using Deep Learning}
In this section, we present our experiments on automatically reconstructing the VMS of an arbitrary image using deep convolutional neural networks (CNN).

\subsection{Image Structures in a Deep CNN}
In order to reconstruct the VMS of an arbitrary image, we utilize the output of a CNN, which is in part transferred and in part trained with a limited number of image-VMS pairs. The purpose of this new deep CNN is to reconstruct the VMS of a given image after training with a given database.	 In this section we analyse the self-emergent image structures that are obtained as outputs of certain neurons in the deep convolution layers of a CNN. Our intention is to analyse the relation between the self-emergent image structures in a CNN and the visual schemas defined by human observers in an image. 

For this purpose, the convolution layers of a deep pre-trained CNN are transferred to a new structure, in which new \emph{fully connected} layers with multiple neurons are added at the output layer and then trained. The aim is to assess whether a CNN, after training the appended layers, is able to reconstruct the combined VMS, {\em i.e.} including both true and false VMSs, for a given image. {While CNNs have been used in various other applications, this study is the first to use them for reconstructing memorable regions of an image}. According to our previous experiments, we can hypothesize that VMSs represent image structures corresponding to semantically distinctive regions in images that can be reconstructed by a deep enough CNN, if the receptive fields of the neurons on each layer are wide enough.

There is still ongoing discussion on how similar and thus transferable are features extracted from images by different deep CNNs \cite{Long2015,Yosinski2015a}. According to the network structure, the optimization method, and the training image set, the internal representations in deep CNNs are expected to be different from layer to layer. {In the following we use transfer learning at various suitable layers in five different CNN architectures, namely} MemNet \cite{Khosla2015}, VGG-S, VGG-M \cite{Chatfield14}, VGG-VD-16L and VGG-VD-19L \cite{Simonyan15}.

\begin{table}[h]
\centering
\begin{tabular}{|l|c|c|}
\hline
\multicolumn{1}{|l|}{\textbf{Input Layer}} & \multicolumn{2}{c|}{224$\times$224$\times$3 RGB Image}\\ \hline
\multicolumn{3}{|c|}{\emph{initial pre-trained (MemNet or VGGs) layers up to a selected L$^{th}$ layer}}\\ \hline \hline\hline
\multicolumn{3}{|c|}{the appended layers}\\ \hline
\emph{layer no. - (type)}  & \emph{weight vector size} & \emph{output blob size}\\ \hline
\textbf{L+1} - (fully con.)& m$\times$n$\times$f$\times$256 & 1$\times$1$\times$256\\ \hline
\textbf{L+2} - (fully con.) & 1$\times$1$\times$256$\times$256 & 1$\times$1$\times$256\\ \hline
\textbf{L+3} - (fully con.) &  1$\times$1$\times$256$\times$400 & 1$\times$1$\times$400\\ 
\hline
\multicolumn{1}{|l|}{\textbf{Output Layer}} & \multicolumn{2}{c|}{400x1 vector (20x20 VMS)}\\ \hline
\end{tabular}
\caption{The generic structure of the CNNs used in the experiments is given. For different pre-trained networks and for different layers selected from these networks, the CNN structures vary.}
\label{TheNet}
\end{table}

MemNet \cite{Khosla2015} is a deep CNN, trained using the output of a large-scale memorability experiment, in which memory scores of 60K images are collected from human observers. The reason we choose this network is to understand whether the image structures that emerge at the layers of MemNet are useful for reconstructing the VMSs obtained in our image memory study. {For this purpose we compare the reconstruction performance of MemNet with four different VGG networks of various depths, which were originally used for category recognition in} \cite{Chatfield14} and \cite{Simonyan15}.

{VGG networks, namely} VGG-S, VGG-M \cite{Chatfield14}, VGG-VD-16L and  VGG-VD-19L \cite{Simonyan15} are four different CNNs of various depths, which are trained with the ImageNet dataset \cite{ILSVRC2015}. {VGGVD-19L, the deepest of them with 19 layers was the winner of the ImageNet, Large Scale Visual Recognition Challenge in 2014. All VGG networks are  composed of varying numbers of convolutional layers succeeded by fully connected layers. VGG type networks are well known in the machine learning community and considered as appropriate for searching for schema-like image structures at their deep layers, because of two reasons. Firstly, the category recognition problem has been shown to create abstract image structures} \cite{Yosinski2015} on the ImageNet dataset. {Secondly, VGGs of different depths would give us a clue about the level of CNN's depth required for reconstructing VMSs.}

\subsection{Deep Learning in Image Memorisation}

{Twenty-one different CNN architectures from five afore-mentioned pre-trained networks are adapted through transfer learning in order to be used for replicating the memory results obtained from humans. The generic structure of the CNNs used in the experiments is given in Table} \ref{TheNet}, where the transferred network is attached to a set of fully connected layers having 256 nodes at each hidden layer and 400 nodes at the output layer. The output layer provides the reconstructed VMS structure as an image of resolution of 20$\times$20 pixels, ensuring a sufficient level of detail. 



Since we use the output of the convolution layer of the transferred CNN as an input to our newly created fully connected head, we produce 21 different CNN structures for our experiments: 3 using each MemNet, VGG-S, VGG-M, VGG-VD-16L (thus a total of 3$\times$4=12), and 9 based on the VGG-VD-19L architecture. As seen in Table \ref{TheNet}\ the initial pre-trained layers upto a selected layer of a network are cut and the neuron outputs are transferred as inputs to our new learning structures. During each separate experiment, a new CNN is created by transferring the layers up to a selected layer, to be trained in order to reconstruct the VMS of an image. The learning rates for the transferred layers are set to zero, so their weights are kept constant during training. The name used for each experiment from this study carries the label of either MemNet or VGG layers which was cut in order to be transferred. For example \emph{experiment conv-5$^2$} indicates that the first 14 convolution layers of VGG-VD-19L network architecture are transferred, while the training takes place for the fully connected layers, as given in Table \ref{TheNet}.

\begin{figure*}[t]
\centering
\begin{subfigure}{14cm}	
	\includegraphics*[trim=+175 +65 +60 +65,clip=false,width=15cm, height=5.7cm]{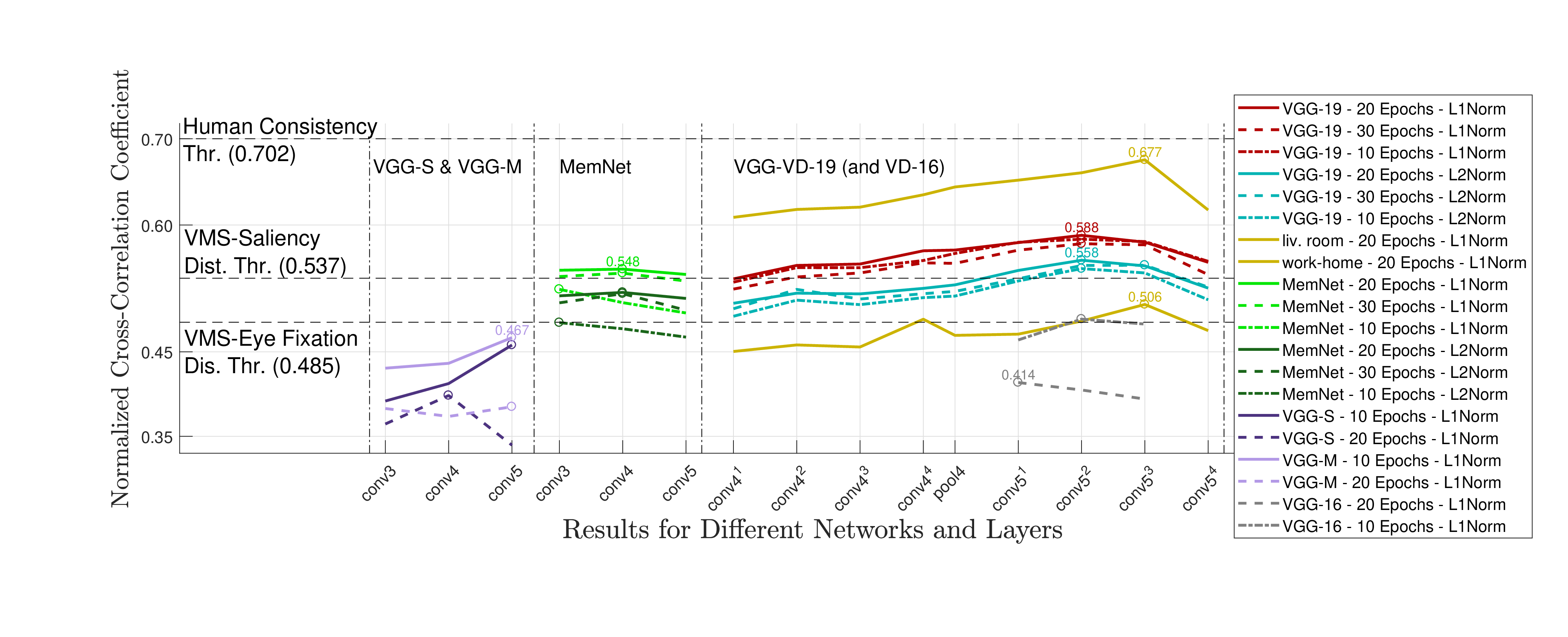}
\end{subfigure}
\caption{VMS reconstruction results when using transfer learning on different layers of the MemNet, VGG-S, VGG-M, VGG-VD-16L and VGG-VD-19L when considering either \emph{l$_1$} or \emph{l$_2$} norms as cost functions, for different numbers of training epochs, when considering the whole VISCHEMA image set and individually for two categories, as specified. The circles on the plots indicate the best performing layer for that particular experiment. The names of the layers are specific to the pre-trained CNN they belong to.}
 \label{VMSResults}
\end{figure*}

\subsubsection{Training and Data Augmentation}
In order to train the CNNs, 80\% of the VISCHEMA image set is used for each experiment. Thus, 640 images, representing 80 images from each category, are used to train the fully connected layers. Each experiment is executed five times, using a different image subset containing 20\% of the whole image set. In the following we consider {21 network structures based on the five pre-trained CNNs, each trained for 5 different runs, when considering 2 different loss functions, leading to a total of 210 experiments.}

In order to enlarge the training set, we implemented a procedure well known for CNNs, called augmentation, by producing mirror images, dividing images into quarters and their mirrors, resulting in a training set which is ten times larger than the initial data. Augmenting by rotating or changing the colour of the images is not used in these experiments because the VMS is a structure, created by human participants, which is susceptible to changes in colour or orientation. The original resolution of the VISCHEMA image set is 700$\times$700 pixels. {Both the training and test images, as well as the augmented image set, are resized to the resolution of} 224$\times$224 {pixels which are then fed into the input layer of the pre-trained networks.}

The VMS maps of the VISCHEMA image set have a resolution of 700$\times$700 pixels. The VMS maps are resized to 20$\times$20 pixel resolution allowing us to reduce the amount of data input that in turn reduces computational complexity required during training. It is possible to do this since VMS's are human annotations that are generally rough and a 20$\times$20 pixel image structures preserves well the VMS signal structure.

The output of the fully-connected network head consists of a vector with 400 components, representing 20$\times$20 pixels image data. Training such as structure corresponds to a multi-dimensional regression problem. To solve this problem, two different loss functions, representing the \emph{l$_1$-norm} and the \emph{l$_2$-norm} are implemented when training the CNNs using back-propagation and stochastic gradient descent with momentum. Batch normalization is used with a batch size of 40 images.
The training\footnote{{Stochastic Gradient Descent (SGD) algorithm with momentum is employed, considering Momentum: 0.9, Initial Learning rate: 0.001 , Weight Decay: 0.0005.}} is performed, using MatConvNet library \cite{MatConvNet}, on a desktop system with dual 2.6Ghz processors and GPU support. Each epoch for an experiment takes approximately 10 minutes for the VGG-VD-based networks and 1 minute for the MemNet and VGG-S/M-based networks. {All 210 experiments are run for 30 epochs, resulting in a total of approximately 25 days of computation.}

\subsection{Reconstruction Results}
The results of each experiment are evaluated by calculating the 2D normalized-cross correlation, {\em i.e.} the Pearson's correlation coefficient: \(\rho\)$^{2D}$, between each reconstructed VMS, representing the emerging VMS calculated by the proposed CNN computational architectures, and the VMSs empirically provided by human observers during the memory experiment. For each experiment we calculate 800 correlation values for the whole VISCHEMA image set.

Figure \ref{VMSResults} shows the reconstruction results for all experiments, when using either l$_1$ or l$_2$ norms as a loss function, after 10, 20 or 30 epochs, for the whole VISCHEMA image set. In order to evaluate the reconstruction results of an experiment, the average correlation between the reconstructed and empirically collected VMSs is compared to the VMS consistency as given in Figure \ref{PartCorr}.b (represented with dashed line on Figure \ref{VMSResults}). This value indicates the upper limit for the memorised image reconstruction based on deep learning architectures.

The l$_1$-norm performs significantly better than l$_2$, when using transfer learning at any of the layers considered. Findings indicate over-fitting occurring after epoch 20 in almost all experiments. The best results are produced by  transfer learning at the Layer-14 (\emph{conv-5$^2$}) of VGG-VD-19L with l$_1$-norm loss function, corresponding to \(\rho\)$^{2D}$=0.588 at epoch 20.  This layer outperforms all other layers in all experiments when the entire VISCHEMA image set is considered. {Deeper layers of VGG-VD-19L perform considerably better in reconstructing the VMS when compared either to the other pre-trained networks, or with the more incipient layers of VGG-VD-19L. MemNet's and VGG-VD-16L's reconstruction success, similarly to VGG-VD-19L's first layers, is comparable to what we obtained when we tested the similarity of the VMS with visual saliency. On the other hand, VGG-S's and VGG-M's reconstruction successes are poor. This indicates that the shallow layers of a CNN, when compared to deeper layers, are unsuccessful in creating the necessary image structures that represent visual memory schemas.}

\begin{figure*}[t]
\centering
\includegraphics*[width=18cm, height=5.3cm]{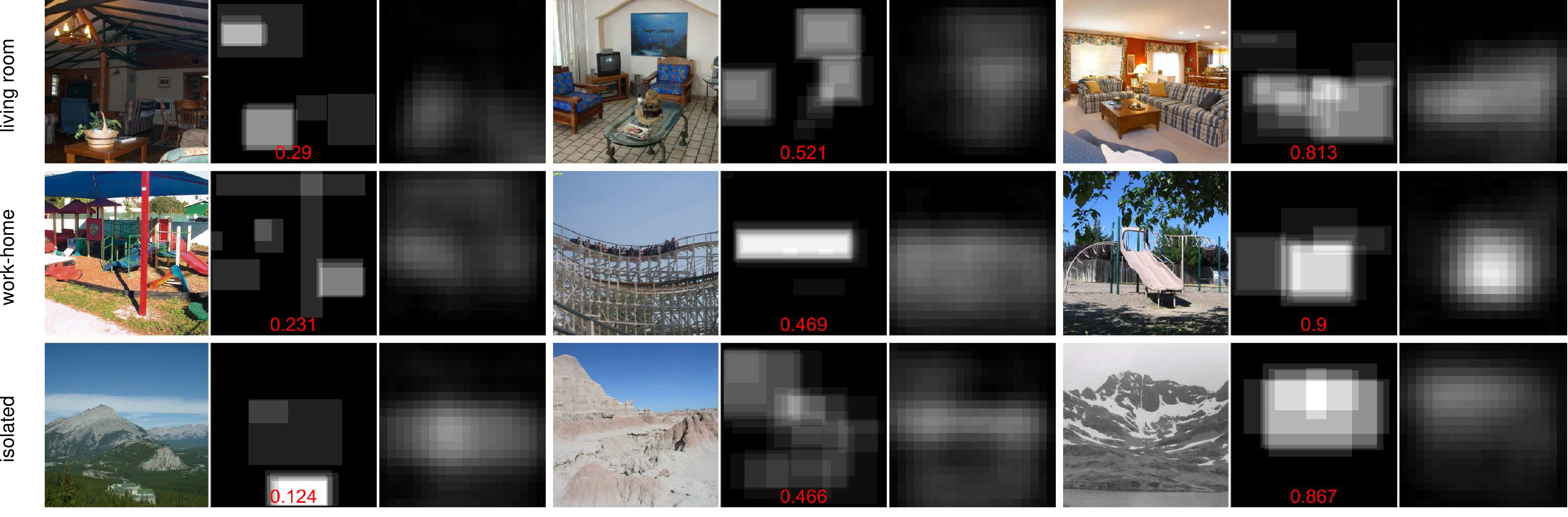}
\caption{Some examples of VMS reconstruction results when using transfer learning for the VGG-VD-19L structure at layer \emph{conv-5$^2$}, using l$_1$-norm, for epoch 20. In each row, there are 3 images for 3 categories, each representing a poor, a moderate and a successful reconstruction result, from left to right, respectively. For each sample image, the images show the empirically collected VMS and that reconstructed using deep learning. The reconstruction accuracy is indicated by the correlation between the empirically collected and reconstructed VMS.}
 \label{SampleResults}
\end{figure*}

In Figure \ref{VMSResults}, we also plot the results of VMS reconstruction using the CNN structure VGG-VD-19L, for two image categories that show the highest and lowest memorability scores, represented by \emph{work-home} and \emph{living room} image categories, respectively. Although there is an exception for the outstanding performance at Layer-15 (\emph{conv-5$^3$}) for the \emph{work-home} category, corresponding to \(\rho\)$^{2D}$=0.677 at epoch 20, the results for structures that emerge at Layer-14 provide the best VMS reconstruction performance across all categories. Some examples when reconstructing VMSs from images, using VGG-VD-19L, are shown in Figure~\ref{SampleResults}.
The most veridical reconstruction of memorable regions for some images is obtained when using transfer learning at certain layers of the VGG-VD-19L, a CNN which was not originally trained \cite{Simonyan15} for image memorability prediction purposes. Although outstanding results are obtained for certain categories, such as for \emph{work-home} category, the reconstruction performance is consistently low for some other categories, such as the \emph{living room} for example. We believe that one reason behind these variations in performance for different image categories is the fact that the image structures in VGG-VD-19L originally emerged for the purpose of category recognition and memorable regions are not necessarily correlated with features characterizing objects used for machine learning recognition. {Texture-like features, used by the VGG-VD-19L network, that are decisive for differentiating the patterns of one cushion cover from another, like the ones we see in the} \emph{living-room} category, are not the ones that can reconstruct a schema of a specific {living-room scene}. This observation is evident in the results from Figure \ref{CatResMemo} for the \emph{living room} category, where VMS pooling did not increase the performance of machine learning predicting image memorability. These results indicate that the Visual Memory Schemas can be reconstructed well for certain categories of images, when using deeper CNNs, despite having a rather small training set.

%

\section{Conclusions}

The main goal of this paper is to characterize image memorability. We introduce the concept of Visual Memory Schema (VMS), and define it as the accumulated memorable parts of an image shared across observers. Visual memory schemas, a concept  derived from the idea of a cognitive schema from Psychology, comprise a mental representation, organization or structure applied to an image, which are shared by observers, allowing us to talk about concepts like the memorability of an image. After conducting a standard episodic memory test on human observers, VMSs were constructed from accumulated human annotations of the memorable regions in each image during the memory experiment. The results show a strong inter-observer correlation for visual memory schemas across all images independent whether they are correctly or incorrectly rated as seen before. This fact suggests that what observers find memorable in images is not only determined by the intrinsic features of images themselves but also by the schemas or mental representations shared by observers about what an image should contain or look like. We show that computational visual saliency and eye fixations are not strongly correlated with what we think that we remember and consequently are poor predictors of image memorability.

Previous studies considered image memorability only as an intrinsic property of the image. In this study we show that memorability of an image is a function of two main factors both embodied in the VMS signal. One factor, known from previous studies, are the intrinsic features of the image, which can be extracted using computer vision algorithms. The other, proposed in this paper, is the collection of visual information structures, shared by human observers, likely to represent the results of their shared experiences and knowledge. {What makes VMS more than just a reformulated intrinsic property of the image is that they are general structures or organizational rules for incoming information employed by human observers that can generalize across images and are not directly tied to a specific image per se.} To this end we also show that shared human experience can be collected via an improved episodic memory experiment, and represented in the form of Visual Memory Schemas. Using both the properties of computer vision features and the shared human visual experience, represented by VMSs, the memorability of an image can be predicted more accurately. 

In a second part of this research study we employed deep learning in order to replicate the results provided by humans during the memory experiment. {Transfer learning was used at various layers on CNN structures such as VGG-VD- 19L, VGG-VD-16L, VGG-S, VGG-M and MemNet. As CNNs get deeper, the features that emerge at their layers become more abstract, conceptual and meaningful. The deep features provided by VGG-VD-19L network lead to significantly better reconstructions of the VMSs in certain image categories, when compared with other VGGs as well as with MemNet, despite the latter being specifically designed for image memorability.} The results are remarkable, given the limitations of the training set, where we were not able to acquire data from thousands of human subjects. The fact that it is these conceptual/abstract layers that better characterise human memory representations than the primitive/signal-based features alone, indicate the limitations of the existing artificial structures in replicating human memory capability.  In order to better understand or predict image memorability one needs to incorporate and account for visual schemas intrinsically shared among human observers.

\bibliographystyle{IEEEtran}
\bibliography{IEEEabvr,memorability}

\vspace{-3.3cm}
\begin{IEEEbiography}[{\includegraphics[trim=25 0 25 0,clip=true,width=1.1in,height=2in,clip,keepaspectratio]{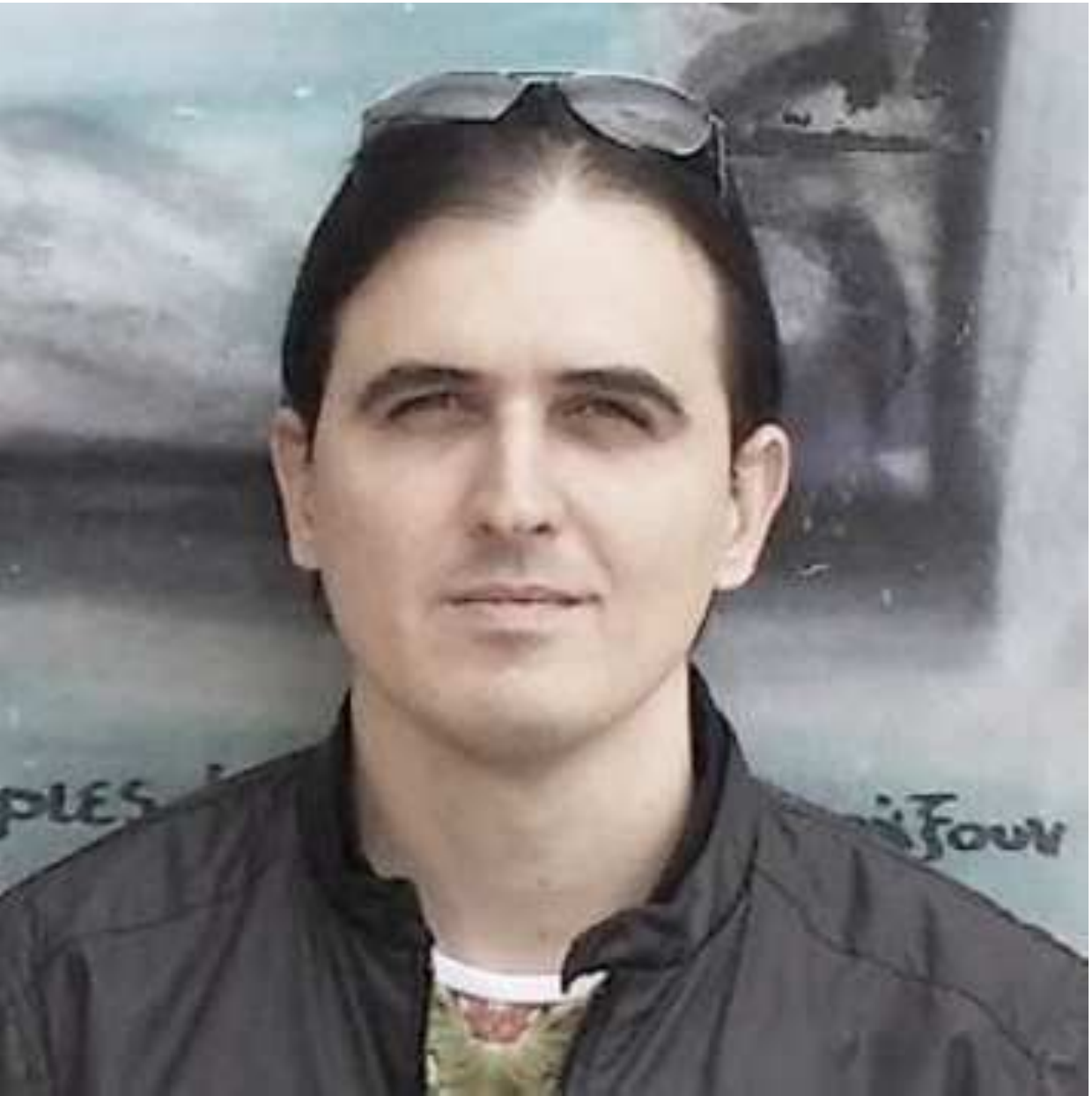}}]{Erdem Akagunduz} is currently an Assistant Professor in Electrical and Electronics Eng. Department, \c{C}ankaya Univ., Turkey. He received the B.S., M.S and Ph.D. degrees in Electronics Engineering from METU, Ankara, Turkey, in 2001, 2004 and 2011, respectively. From 2001 to 2008, he was a research assistant with the METU Computer Vision and Intelligent Systems Laboratory. Between 2009-2016 he worked as a computer vision scientist with ASELSAN Inc. He was a research associate at the University of York, UK, in 2016. His research interests include infra-red computer vision, object/target/scene recognition and tracking.
\end{IEEEbiography}

\vspace{-3.3cm}

\begin{IEEEbiography}[{\includegraphics[trim=0 0 0 0,clip=true,width=1.1in,height=2in,clip,keepaspectratio]{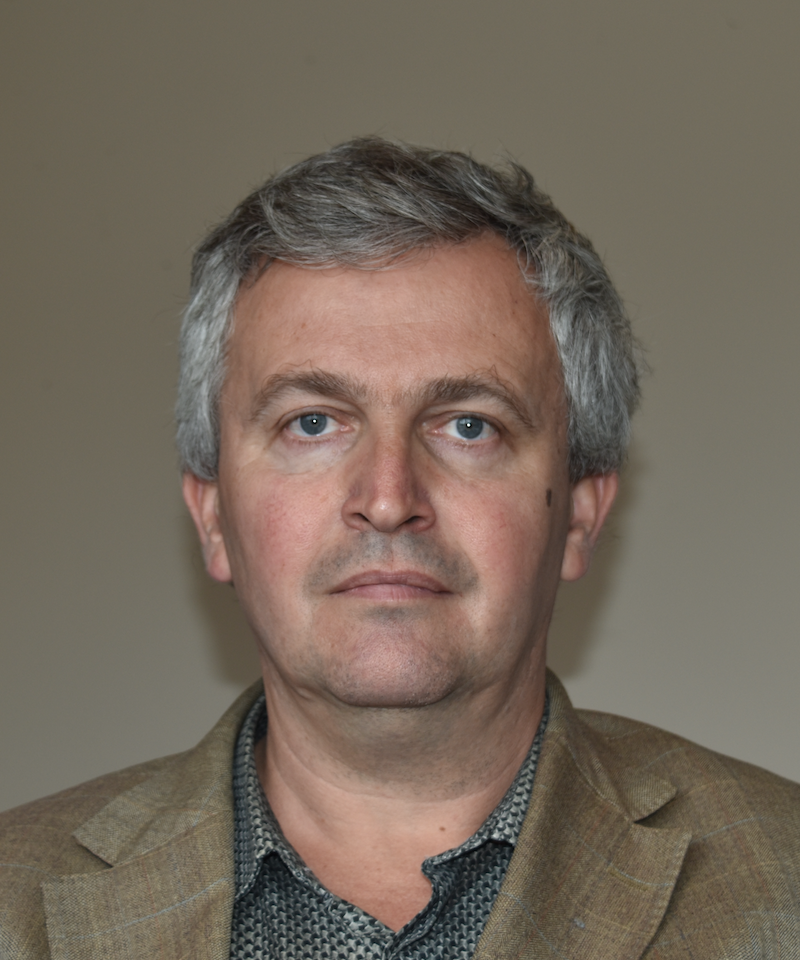}}]{Adrian G. Bors}
is currently a lecturer in Dept. Computer Science, University of York. He received the MSc degree in Electronics Engineering from the Polytechnic Univ. of Bucharest, Romania, in 1992, and the Ph.D. degree in Informatics from the Univ. of Thessaloniki, Greece in 1999. In 1999 he joined the Department of Computer Science, Univ. of York, U.K., where he is currently a lecturer. Dr. Bors was also a Research Scientist at Tampere Univ. of Technology, Finland, a Visiting Scholar at the Univ. of California at San Diego (UCSD), and an Invited Professor at the Univ. of Montpellier II, France. Dr. Bors has authored and co-authored more than 110 research papers including 26 in journals. His research interests include pattern recognition, computer vision and computational intelligence.
Dr. Bors was an associate editor of IEEE Trans. Image Processing between 2010 and 2014 and of IEEE Trans. Neural Networks from 2001 to 2009. He was a co-editor for a special issue for the Journal of Pattern Recognition in 2015. Dr. Bors was a member of the organisation committees for IEEE ICIP 2018, BMVC 2016, CAIP 2013 and IEEE ICIP 2001. He is a Senior Member of the IEEE.
\end{IEEEbiography}

\vspace{-3.3cm}

\begin{IEEEbiography}[{\includegraphics[trim=0 0 0 0,clip=true,width=1.1in,height=2in,clip,keepaspectratio]{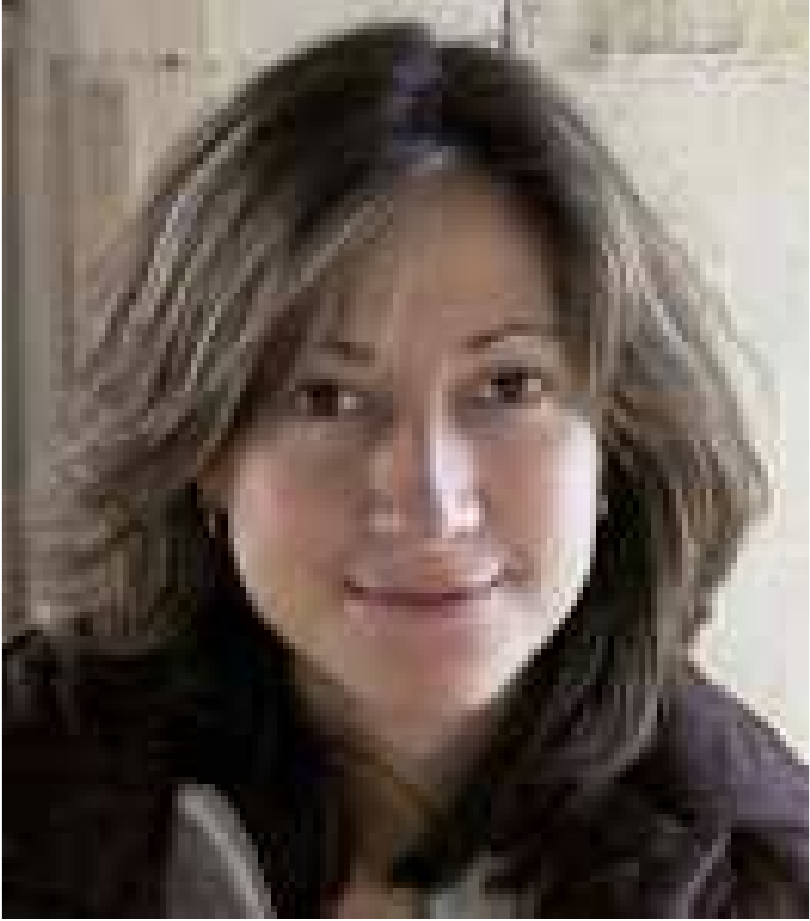}}]{Karla K. Evans}
is currently an Assistant Professor (Lecturer) in the Psychology Department, University of York (UK). She received the PhD degree from Princeton University (U.S.A.) in 2007 and then went on to complete a post-doctoral fellowship at MIT in 2008. Subsequently she worked as a post-doctoral associate at Harvard Medical School and Brigham and Women's Hospital form 2008 to 2013. Her current research interests include visual awareness and visual search, visual episodic memory, perceptual expertise and medical image perception. In addition to leading the Attention and Perception lab at the Psychology Department University of York she is an associate editor for the Journal of Attention, Perception and Psychophysics.
\end{IEEEbiography}

\end{document}